\def\isarxiv{1} 

\ifdefined\isarxiv
\documentclass[11pt]{article}

\usepackage[numbers]{natbib}

\else

\documentclass{article}
\usepackage{neurips_2023}

\fi

\usepackage{amsmath}
\usepackage{amsthm}
\usepackage{amssymb}
\usepackage{algorithm}
\usepackage{subfig}
\usepackage{algpseudocode}
\usepackage{graphicx}
\usepackage{grffile}
\usepackage{wrapfig,epsfig}
\usepackage{url}
\usepackage{xcolor}
\usepackage{epstopdf}
\usepackage{multicol}
\usepackage{bbm}
\usepackage{dsfont}

\allowdisplaybreaks

\ifdefined\isarxiv

\usepackage{tikz}
\usepackage{hyperref}  
\hypersetup{colorlinks=true,citecolor=blue,linkcolor=blue} 
\usetikzlibrary{arrows}
\usepackage[margin=1in]{geometry}

\else

\usepackage{microtype}
\usepackage{hyperref}

\fi
\graphicspath{{./figs/}}

\theoremstyle{plain}
\newtheorem{theorem}{Theorem}[section]

\newtheorem{lemma}[theorem]{Lemma}

\theoremstyle{definition}
\newtheorem{definition}[theorem]{Definition}

\theoremstyle{remark}
\newtheorem{remark}[theorem]{Remark}

\newcommand{\wh}{\widehat}
\newcommand{\wt}{\widetilde}
\newcommand{\ov}{\overline}
\newcommand{\eps}{\epsilon}

\newcommand{\R}{\mathbb{R}}

\renewcommand{\varepsilon}{\epsilon}
\renewcommand{\tilde}{\wt}

\renewcommand{\eps}{\epsilon}
\newcommand{\diag}{\textrm{diag}}
\renewcommand{\d}{\mathrm{d}}

\DeclareMathOperator*{\E}{{\bf {E}}}

\DeclareMathOperator{\poly}{poly}

\title{Federated Empirical Risk Minimization via Second-Order Method}

\ifdefined\isarxiv
\date{}
\author{
Song Bian\thanks{\texttt{BianSongHZ@gmail.com}. University of Wisonsin, Madison.}
\and
Zhao Song\thanks{\texttt{zsong@adobe.com}. Adobe Research.}
\and 
Junze Yin\thanks{\texttt{junze@bu.edu}. Boston University.}
}

\else

\fi

\begin{document}

\ifdefined\isarxiv
\begin{titlepage}
  \maketitle
  \begin{abstract}
Many convex optimization problems with important applications in machine learning are formulated as empirical risk minimization (ERM). There are several examples: linear and logistic regression, LASSO, kernel regression, quantile regression, $p$-norm regression, support vector machines (SVM), and mean-field variational inference. To improve data privacy, federated learning is proposed in machine learning as a framework for training deep learning models on the network edge without sharing data between participating nodes. In this work, we present an interior point method (IPM) to solve a general ERM problem under the federated learning setting. We show that the communication complexity of each iteration of our IPM is $\tilde{O}(d^{3/2})$, where $d$ is the dimension (i.e., number of features) of the dataset.

\end{abstract}
  \thispagestyle{empty}
\end{titlepage}

{\hypersetup{linkcolor=black}
\tableofcontents
}
\newpage

\else

\maketitle
\begin{abstract}
Many convex optimization problems with important applications in machine learning are formulated as empirical risk minimization (ERM). There are several examples: linear and logistic regression, LASSO, kernel regression, quantile regression, $p$-norm regression, support vector machines (SVM), and mean-field variational inference. To improve data privacy, federated learning is proposed in machine learning as a framework for training deep learning models on the network edge without sharing data between participating nodes. In this work, we present an interior point method (IPM) to solve a general ERM problem under the federated learning setting. We show that the communication complexity of each iteration of our IPM is $\tilde{O}(d^{3/2})$, where $d$ is the dimension (i.e., number of features) of the dataset.
\end{abstract}

\fi

\section{Introduction}\label{sec:intro}

Empirical Risk Minimization (ERM) is one of the key problems in machine learning research. ERM appears in many machine learning problems including LASSO~\cite{t96}, logistic regression~\cite{c58,hls13}, support vector machines~\cite{cv95}, 
AdaBoost~\cite{fs97}, kernel regression~\cite{n64,w64}, etc. Due to its wide applications, a great number of works have considered this problem. They not only study the statistical convergence properties but also investigate how to develop efficient algorithms for ERM. Among these efficient algorithms, Interior Point Methods is one of the most widely-used optimization algorithm. IPM is first proposed by~\cite{k84}. After that, IPM has become an active area in optimization research. There is a long line of work using IPM to speedup optimization problems, such as linear programming \cite{cls19,jswz21,sy21}, semi-definite programming \cite{jkl+20}, and cutting plane method \cite{jlsw20}. Recently, \cite{lsz19} develops a fast and robust algorithm to solve Empirical Risk Minimization (ERM).

However, users are not willing to share data with others. Therefore, Federated Learning, which is a general framework for distributed learning on sensitive data, is paid more attention to recently. Motivated by the \textsc{Sketched-SGD}~\cite{irubsa19} and \textsc{FetchSGD}~\cite{rpuisbga20}, there exists a large number of works focus on reducing the communication cost~\cite{jsttkhj14, kmy+16, lhm+17}. In addition, some works~\cite{lsz+20} develop optimization algorithms under federated learning. Nevertheless, all of them develop distributed SGD, which is a first-order optimization algorithm. Due to the reason that first-order algorithms for ERM always depend polynomially on the Lipschitz constant of the gradient and the running time will also have to depend on the strong convexity of the optimization function~\cite{lsz19}. In view of this, we focus on developing distributed second-order optimization algorithms in this paper. As for the distributed second-order optimization algorithm, \cite{ggdkrm21} develops a distributed second-order method, which could address the bottleneck of distributed setting. However, in order to present convergence analysis, they make several strong assumptions that are unrealistic in practice. 

In this work, we mainly study the ERM under FL, we called it FERM (\underline{F}ederated \underline{E}mpirical \underline{R}isk \underline{M}inimization). We develop an IPM framework under FL settings to address FERM first. Then, considering the communication issue of the IPM framework under FL settings, we use sketching techniques to reduce the communication cost. In the end, we present the convergence analysis of our algorithm.

\paragraph{Challenges.} We have witnessed the success of the first-order optimization algorithm under FL. Nevertheless, it is non-trivial to design an IPM under FL. Especially, we need to use the sketching technique to reduce the communication cost and provide convergence analysis for IPM under FL. In the following sections, we focus on answering the following problems: 
\begin{itemize}
    \item \textit{How to design a distributed IPM algorithm without data sharing?}
    \item \textit{How to use sketch matrices to compress the Hessian information under distributed setting?}
    \item \textit{Is it possible to present convergence guarantees of IPM under FL?}
\end{itemize}

Before we show the specific algorithms and analysis, we first state our main result here:
\begin{theorem}[Informal Main Result, see Appendix~\ref{sec:proof_main_res} for details] \label{thm:main_result_informal}
If the following conditions hold
\begin{itemize}
    \item Consider a convex problem under the FL setting $\min_{Ax=b, x \in \Pi_{i=1}^{m}K_{i}}c^{\top}x$, where $K_{i}$ is compact convex sets.
    \item For each $i \in [m]$, we are given a $\nu_{i}$-self concordant barrier function $\phi_{i}$ for $K_{i}$. 
\end{itemize}

Then, there exists a FL algorithm (see Algorithm~\ref{alg:erm_fl}) runs in $O(\sqrt{\nu} \log^{2} m \log(\frac{\nu}{\delta}))$ iterations and each iteration sends $O(b_{\max}n)$ bits to find a vector $x$ up to $\delta$ error, where $b_{\max}$ is determined by the size of sketch matrices.
\end{theorem}

\paragraph{Contributions.} Our contributions are summarized as follows: 
\begin{itemize}
    \item To the best of our knowledge, we first study the ERM under FL settings. And we propose an IPM under FL to solve FERM.
    \item We are also the first one to sketch Hessian information of the IPM algorithm under the FL setting. Previous works only either sketch the gradient information under the FL setting or Hessian information under the classical distributed computing setting.
    \item We show convergence guarantees of IPM under FL, which compresses the Hessian via sketching methods to reduce communication costs. Due to the reason that IPM is a second-order optimization method, it is non-trivial for us to present such convergence without making strong assumptions.
\end{itemize}

\paragraph{Organization.} We present related work in Section~\ref{sec:related_work}. In Section~\ref{sec:background}, we present the background of this paper. And in Section~\ref{sec:prob_form}, we formulate the problem first. Next, we give the sketching technique we used in our algorithm and the overview of our main algorithm. 
In Section~\ref{sec:theo}, we present the theoretical analysis of our algorithm. In Section~\ref{sec:comparison}, we compare our algorithm with some naive models. And we conclude this paper in Section~\ref{sec:discussion}.

\section{Related Work} \label{sec:related_work}

\paragraph{Distributed Optimization Methods.} 

Nowadays, distributed optimization methods have gained popularity. As for the distributed first-order optimization methods, a large number of works focus on developing communication-efficient distributed SGD. These work includes that distributed variants of stochastic gradient descent~\cite{zwsl10,nrrw11,mksb13}, accelerated SGD~\cite{ss14}, variance reduction SGD~\cite{llmy17,rhsps15}, dual coordinate ascent algorithms~\cite{y13,rt16,zwxxz17}, and stochastic coordiante descent methods~\cite{fr16}. As for the distributed second-order optimization methods, DANE~\cite{ssz14}, AIDE~\cite{rkrps16}, and DiSCO~\cite{zl15} are well-known work. CoCoA~\cite{jsttkhj14,msjjrt15,sfctjj18} is similar to the second-order method, but it does not use any second-order information.

\paragraph{Federated Learning.} 
Federated learning is a special case of distributed machine learning. Federated learning allows clients to train machine learning models without data sharing. The applications of federated learning include healthcare~\cite{lgdsvd20,rhlmrabglm20}, financial area~\cite{yzylx19}, and autonomous vehicle~\cite{llcly19}. Although federated learning has numerous advantages, the federated learning is always limited by the communication issue. In view of this, a great number of methods~\cite{irubsa19,rpuisbga20} are developed to reduce communication cost in federated learning. FEDAVG~\cite{mmrha17} is the first work focus on solving communication efficiency problem in federated learning. After that, a great number of gradient compression methods~\cite{irubsa19,rpuisbga20} have been proposed. Communication-efficient algorithms achieve success in practice. However, it is not easy to present convergence analysis for communication-efficient algorithms. Recently, \cite{rpuisbga20} presents convergence analysis for SGD under federated learning. Federated learning convergence on one-layer neural networks is investigated in~\cite{ljzkd21}. Furthermore, \cite{hlsy21} gives convergence guarantees of the general federated learning on neural networks. \cite{swyz23} studies federated learning for convex, Lipschitz, and smooth functions. \cite{lsy23} proposes an federated algorithm for adversarial training in deep learning. Another interesting angle of federated learning is differential privacy, a number of works \cite{hsla20,hsc+20,clsz21,cstz22} have studied the privacy inspired question related to federated learning. Privacy is not the major task in this paper.

\paragraph{Sketching Technique.} 
Sketching technique has been widely applied to many applications in machine learning, such as low-rank approximation~\cite{cw13,nn13,mm13,bw14,swz17,alszz18}, linear regression, distributed problems~\cite{wz16,bwz16}, reinforcement learning~\cite{wzd+20}, tensor decomposition~\cite{swz19_soda}, sparsification of attention matrix \cite{dms23}, discrepancy minimization \cite{dsw22}, clustering~\cite{emz21}, online bipartite matching \cite{swy23,hst+22}, exponential and softmax regression \cite{lsz23,dls23,lsx+23,gsy23_hyper}, integral optimization \cite{jlsz23}, submodular problem \cite{qsw23}, generative adversarial networks~\cite{xzz18}, symmetric norm estimation \cite{dswz22}, optimizing neural tangent kernel \cite{bpsw21,syz21,szz21,hswz22,z22,gqsw22,als+22}, database \cite{qjs+22}, fast attention computation \cite{as23}, dynamic kernel computation \cite{qrs+22,djs+22}, matrix completion \cite{gsyz23}, matrix sensing \cite{dls23_sensing,qsz23}. Count Sketch~\cite{ccf02} is used in~\cite{irubsa19,rpuisbga20} to reduce the cost of communication at each iteration. Count Sketch is able to approximate every coordinate of a vector with an $\ell_{2}$ guarantee. And it is also possible to recover an approximate vector from Count Sketch. In this paper, we use AMS matrices~\cite{ams99} to compress the model updates at each iteration to reduce the communication cost of FL.

\section{Background} \label{sec:background}

In Section~\ref{sub:background:notation}, we explain the notations that we use. In Section~\ref{sub:background:minimization}, we introduce empirical risk minimization. In Section~\ref{sub:background:central}, we explain the central path method and the properties of the self-concordant barrier function. In Section~\ref{sub:background:newton}, we present the Newton method.

\subsection{Notations}
\label{sub:background:notation}

Given a positive value $n$, we use $[n]$ to denote $\{1,2, \cdots, n\}$. We use $m$ to denote the number of clients. For each client $i \in [m]$, it contains dataset $A_{i} \in \R^{d \times n_{i}}$. We also assume that $\sum_{i=1}^{m} n_{i} = n$. Moreover, we define $x$ as the main variable, and $s$ as the slack variable. We use $x_{i}$, $s_{i}$, $W_{i}$ and $A_{i}$ to denote the variables that are computed in client $c_{i}$. And we use $x^{t}_{i}$, $s^{t}_{i}$, $W^{t}_{i}$, and $A^{t}_{i}$ to denote the variables that are computed in client $c_{i}$ at $t$-th iteration.

Next, we define two operations here. The operation $\oplus$ denotes concatenation operation, which indicates that $x = \oplus_{i=1}^{m} x_{i} = [x_{1}, x_{2}, \dots, x_{m}]^{\top}$ and $s = \oplus_{i=1}^{m} s_{i} = [s_{1}, s_{2}, \dots, s_{m}]^{\top}$. And we use $\otimes$ to denote the following operation: $W = \otimes_{i=1}^{m} W_{i} = \diag(W_1, W_2, \cdots, W_m)$.

Given that $f$ and $g$ are two functions, $f \lesssim g$ means that $f \leq Cg$, where $C$ is a constant. Let $v$ be a vector. $\|v\|$ represents the standard Euclidean norm. $\E[]$ represents the expectation and $\Pr[]$ denotes the probability. We use $\nabla f(x)$ to denote the gradient of $f$, namely $\frac{\d f}{\d x}$.

For any $A \in \R^{m \times n}$, $\|A\|_{2}$ represents its operator norm and $\|A\|_{F}$ stands for its Frobenius norm. We also use some facts that $\|AB\|_{2} \leq \|A\|_{2} \cdot \|B\|_{2}$, $\|A\|_{F} \leq \sqrt{n} \|A\|_{2}$. Moreover, if the matrix $A \in \R^{n \times n}$ is a block diagonal matrix, then $A$ could be expressed as $\diag (A_1, A_2, \cdots, A_m)$, 
where $A_1$ is a matrix whose dimensions is $n_1 \times n_1$, $A_2$ is a matrix whose dimensions is $n_2 \times n_2$, and $A_m$ is a matrix whose dimensions is $n_m \times n_m$. In addition, $\sum_{i=1}^{m} n_{i} = n$. If $A \in \R^{n \times n}$ is a symmetric positive semi-definite (PSD) matrix, i.e., $A \succeq 0$ if for all vectors $x \in \R^n$, then $x^\top A x \geq 0$, and we use $\|v\|_{A}$ to denote $(v^{\top}Av)^{1/2}$. If we are given a convex function $f$, we use $\|v\|_{x}$ to denote $\|v\|_{\nabla^{2}f(x)}$ and $\|v\|_{x}^{*}$ to denote $\|v\|_{\nabla^{2}f(x)^{-1}}$ for simplicity.

In general, we use $R \in \R^{b \times d}$ or $S \in \R^{b \times d}$ to denote sketches that are used to compress model updates. In order to distinguish different sketches, we use $R_{i} \in \R^{b_{i} \times d}$ and $S_{i} \in \R^{b_{i} \times d}$.

Furthermore, in this paper, we consider the computation model to be word RAM model. In this model, each word has $O(\log n)$ bits and all the basic computation can be done in $O(1)$. This is standard in the literature of algorithm design~\cite{clrs09} and distributed algorithm~\cite{wz16,bwz16}.

\subsection{Empirical Risk Minimization}
\label{sub:background:minimization}

We give the definition of traditional Empirical Risk Minimization (ERM) as below:
\begin{definition}[Empirical Risk Minimization] Given a convex function $f_i : \R^{d} \rightarrow \R$, $A_{i} \in \R^{d \times n_{i}}$, $x_{i} \in \R^{n_{i}}$ and $b_i \in \R^{d}$, $\forall i \in [m]$, we call the following optimization problem as Empirical Risk Minimization problem $\min_{x} \sum_{i=1}^{m} f_{i} (A_{i} x_{i} + b_{i})$.
\end{definition}
Then, we could rewrite the original problem by defining $y_i = A_{i} x_{i} + b_{i}$, $z_i = f_i( A_{i} x_{i} + b_{i} )$. After that, we could get the following problem: 

\begin{align*}
    \min_{x,y,z} &~ \sum_{i=1}^{m} z_{i} \\
    \mathrm{s.t.} &~ Ax + b = y \\
    (y_{i}, z_{i}) \in K_{i} &~= \{(y_{i}, z_{i}): f_{i}(y_{i}) \leq z_{i}\}, \forall i \in [m]
\end{align*}

In this paper, we mainly consider the following question, when the dimension of $K_i$ could be arbitrary: $\min_{ x \in \prod_{i=1}^{m} K_i, A x = b } c^\top x $. 

In the next section, 
we briefly introduce the solutions to address the general form under centralized setting.

\subsection{Central Path Method}
\label{sub:background:central}

\begin{figure}[!ht]
\centering
\includegraphics[width=0.45\textwidth]{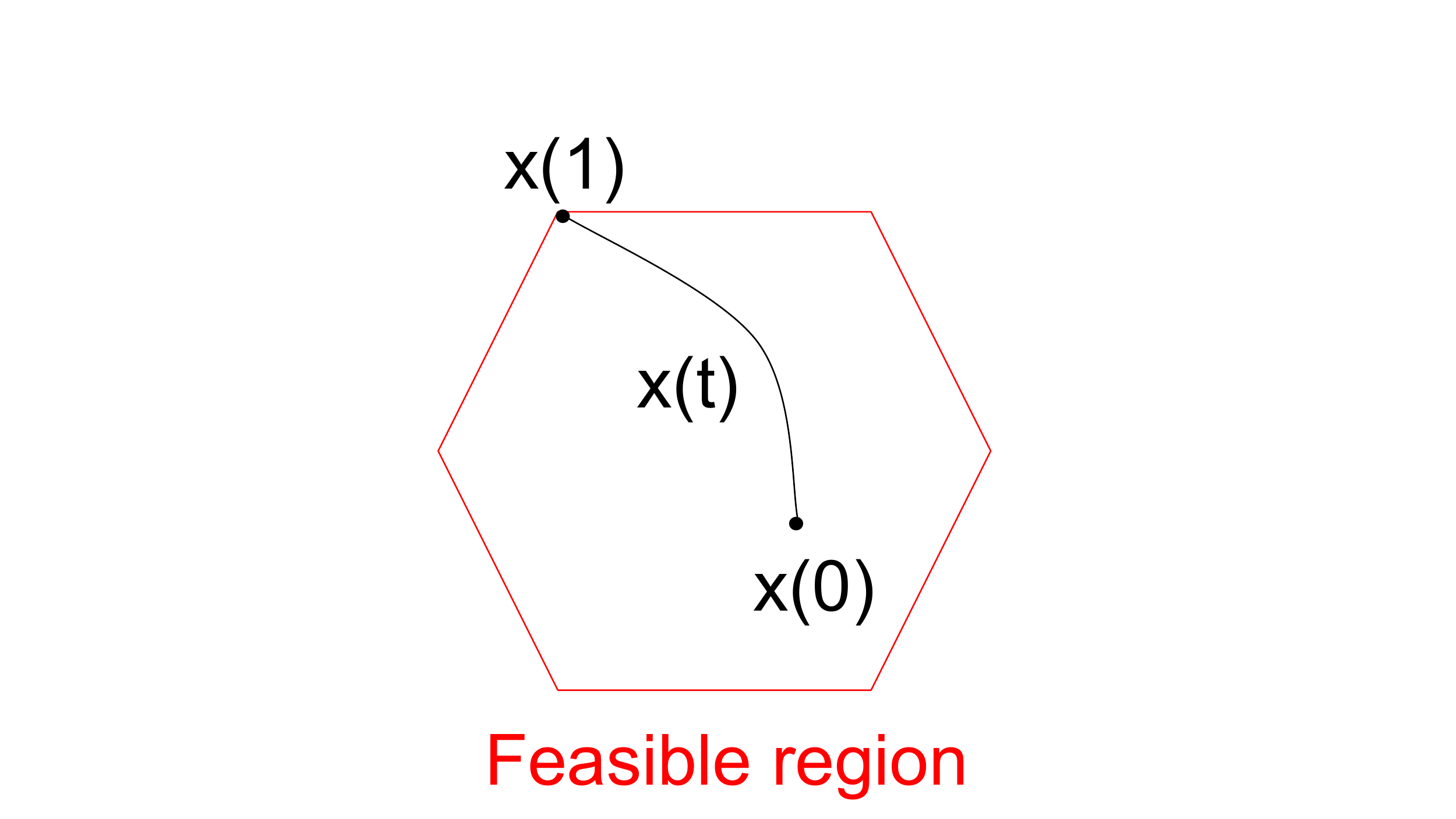} 
\caption{
Here is an example of the central path. The curve denotes the central path. The hexagon denotes the feasible region. The start point of the central path is $x(1)$, and the end point of the central path is $x(0)$. Then we follow the path $x(t)$ to from $x(1)$ to $x(0)$. 
}
\label{fig:central_path}
\end{figure}

In this section, we introduce the central path method. First, we recap the problem that we analyze:
\begin{equation}
\min_{x \in \prod_{i=1}^{m} K_{i}, Ax=b} c^{\top} x
\label{eq:origin_problem}
\end{equation}
For each $i \in [m]$, $K_{i}$ is a convex set and $x_{i}$ is the $i$-th block of $x$ respect to $K_{i}$. The interior point methods (IPM) consider the following path of solutions:
\begin{equation}
x(t) = \arg \min_{Ax=b} c^{\top}x + t \sum_{i=1}^{m} \phi_{i}(x_{i})
\label{eq:central_path}
\end{equation}
where $\phi_{i}(\cdot)$ is called self-concordant barrier function. (Fig.~\ref{fig:barrier_function} is an example of barrier function). And the path is always called central path. The IPM solves Eq.~\eqref{eq:origin_problem} by decreasing $t \rightarrow 0$ (See Fig.~\ref{fig:central_path}). 

The running time of the central path based algorithm is determined by the self-concordant barrier function. In view of this, we first present the definition and properties of self-concordant barrier function here.

\begin{definition}\label{def:phi}
Given a function $\phi$, if any $x\in\mathrm{dom}\phi$ and any $u\in\R^{n}$, the following inequality holds
$
    |\nabla^{3}\phi(x)[u,u,u]| ~\leq 2\|u\|_{x}^{3/2} , 
    \|\nabla\phi(x)\|_{x}^{*} ~\leq \sqrt{\nu}
$
where $\|v\|_{x}:=\|v\|_{\nabla^{2}\phi(x)}$ and $\|v\|_{x}^{*}:=\|v\|_{\nabla^{2}\phi(x)^{-1}}$, for any vector $v$. Then, the function $\phi$ is called as a $\nu$ self-concordant barrier for $K$, where $K=\mathrm{dom}\phi$.
\end{definition}
\begin{remark}
In general, $\nu \geq 1$ for any self-concordant barrier function.
\end{remark}

\cite{n98} demonstrates that for any open convex set $K$ contained in the Euclidean space $\R^{n}$, there exists a $O(n)$ self-concordant barrier function. We focus on a specific convex set $K_{i}$ which has a dimension of $O(1)$ in this paper. We make the assumption that a $\nu_{i}$ self-concordant barrier function $\phi_{i}$ is given, and we can efficiently compute its gradient $\nabla\phi_{i}$ and Hessian $\nabla^{2}\phi_{i}$ in constant time ($O(1)$). An important result we rely on regarding self-concordance is the stability of the norm $\|\cdot\|_{x}$ when we alter the value of $x$. Subsequently, we proceed to present certain properties of the self-concordant barrier function.

\begin{theorem}[Theorem 4.1.6 in \cite{n98}]\label{thm:hessiansc}
If the following conditions hold
\begin{itemize}
    \item Suppose $\phi$ represents a self-concordant barrier function.
    \item the norm $\|y-x\|_{x}$ is less than $1$
\end{itemize}

Then, the following inequalities hold true:
$
\nabla^{2}\phi(y) \succeq (1-\|y-x\|_{x})^{2}\nabla^{2}\phi(x)
$
and
$
\nabla^{2}\phi(y) \preceq{(1-\|y-x\|_{x})^{-2}}\nabla^{2}\phi(x)
$.
\end{theorem}
Now, we consider the way to go alone with the path from $x(1)$ to $x(\eps)$, where $\epsilon \in (0,1)$, in the next section.

\subsection{Newton Step}
\label{sub:background:newton}

In this section, we briefly introduce the Newton method in central path. It is a standard method, for details of the background, the readers could refer~\cite{nw06}.

In order to follow the path from $x(1)$ to $x(\eps)$ and control in error that caused in the progress, we consider the following problem
\begin{align*}
    s/t + \nabla \phi(x) &~= \mu \\ 
    Ax &~= b \\ 
    A^{\top}y + s &~= t \mu + c
\end{align*}
where $\nabla \phi(x) = (\nabla \phi_{1}(x_{1}), \phi_{2}(x_{2}), \cdots, \nabla \phi_{m} (x_{m}) )$ and $\mu$ stands for the error that is caused in the progress. In order to control the error, the Newton step to move from $\mu$ to $\mu + h$ is given below:
\begin{align*}
    \delta^{*}_{s} / t + \nabla^{2} \phi(x) \cdot \delta^{*}_{x} &~= h \\ 
    A \delta^{*}_{x} &~= 0 \\ 
    A^{\top} \delta^{*}_{y} + \delta^{*}_{s} &~= 0
\end{align*}
where $\nabla^{2} \phi(x) = \diag(\nabla^{2} \phi_{1}(x_{1}), \nabla^{2} \phi_{2}(x_{2}), \cdots, \nabla^{2} \phi_{m}(x_{m}))$.

Then, we define that $W := (\nabla^{2} \phi(x))^{-1}$ and we define the projection matrix $P \in \R^{n \times n}$ below: 
\begin{equation} \label{eq:def_P}
P:= W^{1/2}A^{\top}(AWA^{\top})^{-1}AW^{1/2}    
\end{equation} 
We could get the following solutions:
\begin{align*}
    \delta^{*}_{x} ~= W^{1/2}(I-P)W^{1/2}h, ~~~
    \delta^{*}_{y} ~= - t \cdot (AWA^{\top})^{-1}AWh, ~~~
    \delta^{*}_{s} ~= t \cdot W^{-1/2}PW^{1/2}h 
\end{align*}

\section{IPM under FL} \label{sec:prob_form}

In this section, we develop the interior point methods under FL. Before we introduce our algorithm, we first introduce the sketching technique used in section~\ref{sec:sketch_tech}. Then, we give the overview of our algorithm in section~\ref{sec:our_alg}.

\subsection{Sketching Techniques} \label{sec:sketch_tech}

In this subsection, we give the definition of AMS matrix~\cite{ams99} and show the statistical properties of using the AMS matrix to sketch a fixed vector. See Appendix~\ref{sec:fastjl} for rigorous proof.

\begin{definition}[AMS sketch matrices~\cite{ams99}] \label{def:ams_matrix}
Let $h_{1}, h_{2}, \dots, h_{b}$ be $b$ random hash functions. The hash functions are picked from a $4$-wise independent hash family $\mathcal{H} = \{h:[n] \rightarrow \{-1/\sqrt{b}, +1/\sqrt{b}\}\}$. Then $R \in \R^{b \times n}$ is an AMS sketch matrix if we set $R_{i,j}=h_{i}(j)$.
\end{definition}

The AMS matrix has great statistical properties to sketch a fixed vector. We provide the statement in the following lemma, which is standard in literature \cite{lsz19,sy21}.

\begin{lemma}[Statistical properties for sketching a fixed vector]
If the following conditions hold
\begin{itemize}
    \item $h \in \R^n$ is a fixed vector.
    \item $R$ is defined as in Definition~\ref{def:ams_matrix}.
\end{itemize}

Then we have 
\begin{align*}
& ~ \E[ R^\top R h ] = h, ~~~ 
\E[ (R^\top R h)_i^2 ] \leq h_i^2 + \frac{1}{b} \| h \|_2^2 \\
& ~ \Pr \left[ | ( R^\top R h )_i - h_i | > \| h \|_2 \frac{\log ( n / \delta ) }{ \sqrt{b} } \right] \leq \delta
\end{align*}
\end{lemma}

\begin{figure*}[!t]
    \centering
    \includegraphics[width=0.95\textwidth]{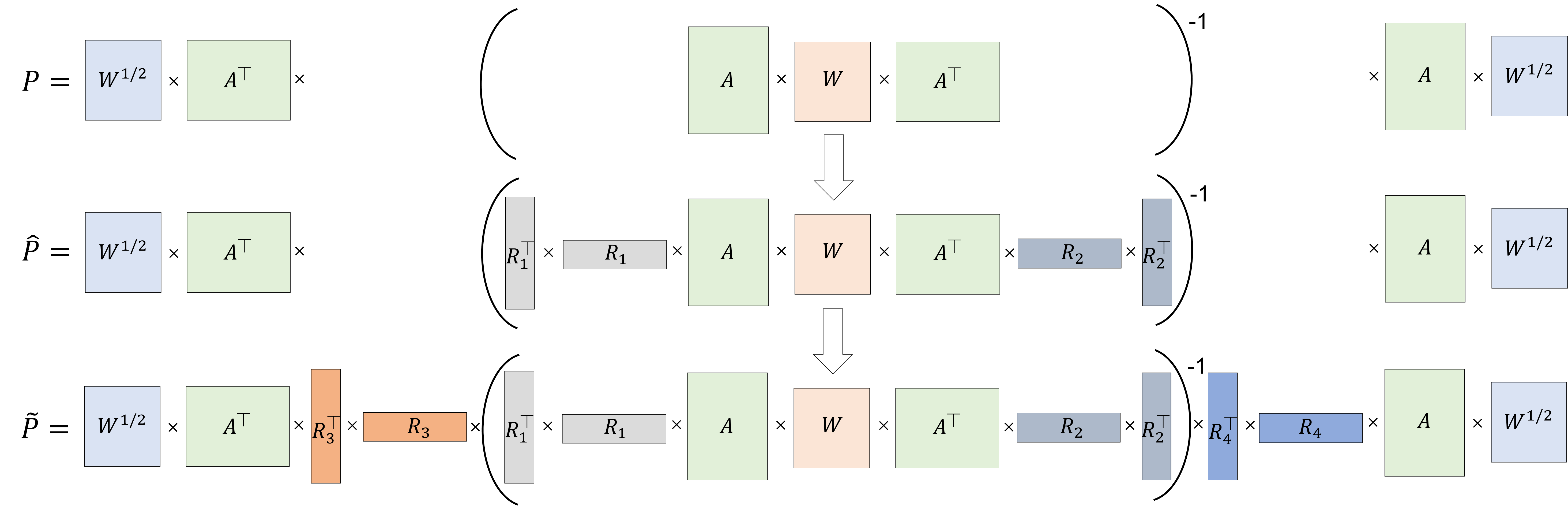}
    \caption{ $P$ is an ideal case of the projection matrix. However, it is infeasible to construct $P$ under FL. In view of this, we construct $\wt{P}$. In order to analyze that $P$ is close to $\wt{P}$, which means that $\|g^{\top} P h - g^{\top} \wt{P} h\|$ could be bounded by $g$, $h$, $A$, and $W$. We create an artificial matrix $\wh{P}$. Note that $\wh{P}$ is only used in analysis. The way how our analysis working is, we first show that $P$ is close $\wh{P}$, then we show $\wh{P}$ is close to $\wt{P}$. Combining two steps together, we finally prove that $P$ is close to $\wt{P}$.
    }
    \label{fig:P_wt_P_wh_P}
\end{figure*}

\textbf{Note.} Although AMS sketch matrix is also used in~\cite{lsz19}, there exists some difference between our paper and the previous work: Both of our work and that use AMS sketch matrix. However, the previous work uses the AMS sketch matrix outside of the projection matrix $P$ to accelerated the whole process, where $R \in \R^{b \times n}$ is an AMS matrix and $P \in \R^{n \times n}$ is a projection matrix. And we add the AMS sketch matrix inside the projection matrix. The projection matrix with AMS sketch matrix in our paper is defined in Def.~\ref{def:app_P}. There is a major issue we need to tackle:  how to bound the error that is caused by adding sketching matrices outside the inverse part ($AWA^{\top}$)?

\subsection{Our Algorithm} \label{sec:our_alg}
In view of the properties of AMS sketch matrices, we could use AMS sketch matrices to bound the error caused by the sketching techniques. 

Next, we define the following notations to differentiate the projection matrices used in IPM:

\begin{definition}[$\wh{P}$ and $\wt{P}$]\label{def:app_P}
Given four independent AMS matrices, $R_1 \in \R^{b_{1} \times d}$, $R_2 \in \R^{b_{2} \times d}$, $R_3 \in \R^{b_{3} \times d}$, $R_4 \in \R^{b_{4} \times d}$, the matrix $\wh{P}$ and $\wt{P}$ are defined as below:
\begin{align*}
\wh{P} = & ~ W^{1/2}A^{\top}(R_1^\top R_1 AWA^{\top} R_2^\top R_2 )^{-1}AW^{1/2} 
\end{align*}
and
\begin{align*}
\wt{P} = & ~ W^{1/2} A^\top R_3^\top R_3 ( R_1^\top R_1 A W A^\top R_2^\top R_2 )^{-1} R_4^\top R_4 A W^{1/2} 
\end{align*}

\end{definition}

We want to remark that $\wh{P}$ is only being used for the purpose of analysis. But, we use $\wt{P}$ for both analysis and algorithm.

The algorithm to address ERM under FL could be divided into several steps (Fig.~\ref{fig:overview} gives an overview of the algorithm):

\begin{algorithm*}[!ht]
\caption{Second-Order Algorithm for Empirical Risk Minimization in Federated Learning (Theorem~\ref{thm:main_result_informal})}
\label{alg:erm_fl}
\begin{algorithmic}[1]
\State Both Server and Client have the random seed for sketching matrix $R_{1}$, $R_{2}$, $R_{3}$, and $R_{4}$.
\State Initial $\delta_{x}^{0} \leftarrow 0$, $\delta_{s}^{0} \leftarrow 0$, $\lambda \leftarrow 2^{16} \log(m)$, $\alpha \leftarrow 2^{-20} \lambda^{-2}$, $\xi \leftarrow 2^{-10} \alpha$, $\wt{t} \leftarrow 1$.
\For{$t = 1 \to T$}
    \State {\color{blue}/* Client */}
    \For{$i = 1 \to N$}
        \State  {\color{blue}/* Update local parameters for $x$ and $s$ */}
        \State $x^{t}_{i} \leftarrow x^{t-1}_{i}+\delta_{x_{i}}^{t-1}$. 
        \State $s^{t}_{i} \leftarrow s^{t-1}_{i}+\delta_{s_{i}}^{t-1}$.
        \State $W^{t}_{i} \leftarrow (\nabla^{2}\phi_{i}(x_{i}^{t}))^{-1}$.
        \State Compute $\mu_{i}^{t}(x,s) \leftarrow s_{i} / \wt{t} + \nabla\phi_{i}(x^{t}_{i})$
        \State Compute $\gamma_{i}^{t}(x, s) \leftarrow \|\mu_{i}^{t}(x,s)\|_{\nabla^{2}\phi_{i}(x^{t}_{i})^{-1}}$
        \State Compute $c_{i}^{t}(x,s) \leftarrow \begin{cases}
        \frac{\exp(\lambda\gamma_{i}^{t}(x,s))/\gamma_{i}^{t}(x,s)}{(\sum_{i=1}^{m}\exp(2\lambda\gamma_{i}^{t}(x,s)))^{1/2}} & \text{if }\gamma_{i}^{t}(x,s) \geq 96\sqrt{\alpha}\\
        0 & \text{otherwise}
        \end{cases}$ 
        \State Compute $h_{i}^{t} \leftarrow -\alpha\cdot c_{i}^{t}(x, s)\mu_{i}^{t}(x,s)$
        \State Send $(W_{i}^{t})^{1/2} A_{i}^{\top} R_{1}^{\top}$, $R_{2} A_{i} W_{i}^{t} A_{i}^{\top} R_{3}^{\top}$, $R_{4} A_{i} (W_{i}^{t})^{1/2}$ and $h_{i}^{t}$ to server.
    \EndFor
    \State {\color{blue}/* Server */}
    \State Construct $h^{t} \leftarrow \oplus_{i=1}^{m} h_{i}^{t}$ \Comment{$h^{t} \in \R^{n}$} 
    \State Construct $\wt{P}^{t} \leftarrow (W^{t})^{1/2}A^{\top} R_{1}^{\top} R_{1} (R^{\top}_{2} R_{2} A W^{t} A^{\top} R_{3}^{\top} R_{3})^{-1} R^{\top}_{4} R_{4}A(W^{t})^{1/2}$ \Comment{ $\wt{P} \in \R^{n \times n}$ }  
    \State Compute $\delta_{x}^{t} \leftarrow (W^{t})^{1/2}(I-\wt{P}^{t})(W^{t})^{1/2}h^{t}$. \Comment{$\delta_{x}^{t} \in \R^{n}$}
    \State Compute $\delta_{s}^{t} \leftarrow \wt{t} \cdot (W^{t})^{-1/2}\wt{P}^{t}(W^{t})^{1/2}h^{t}$. \Comment{$\delta_{s}^{t} \in \R^{n}$}
    \State Send $\delta_{x}^{t}$ and $\delta_{s}^{t}$ to every client.
    \State $\wt{t} = (1 - \frac{\xi}{\sqrt{\nu}}) \wt{t}$.
\EndFor
\State Return an approximation solution to the convex problem. 
\end{algorithmic}
\end{algorithm*}

\paragraph{Setup.} First, we give server and each client the same random seed. Then, for each client $c_{i}$, the client generates four independent sketching matrices $R_{1}$, $R_{2}$, $R_{3}$, and $R_{4}$.

\paragraph{Local update.} For any $i \in [m]$, the detailed process of local update in each client $c_{i}$ is shown as below:
\begin{itemize}
    \item Each client $c_{i}$ updates $x^{t-1}_{i}$ and $s^{t-1}_{i}$ by gradient descent $\delta_{x_{i}}^{t-1}$ and $\delta_{s_{i}}^{t-1}$ respectively. Then, we get that $x^{t}_{i}=x^{t-1}_{i} + \delta_{x_{i}}^{t-1}$ and $s^{t}_{i}=s^{t-1}_{i}+\delta_{s_{i}}^{t-1}$.
    \item Each client $c_{i}$ computes $W^{t}_{i}$.
    \item Each client $c_{i}$ computes $\mu^{t}_{i}(x, s)=s^{t}_{i}/\wt{t} + \nabla \phi_{i}(x_{i})$ and $\gamma^{t}_{i}(x, s) = \| \mu^{t}_{i}(x, s) \|_{\nabla^{2} \phi_{i}(x^{t}_{i})^{-1}}$.
    \item Each client $c_{i}$ computes $h_{i}^{t} = - \alpha \cdot c_{i}^{t}(x, s) \mu_{i}^{t}(x, s)$.
    \item Each client $c_{i}$ sends its $(W_{i}^{t})^{1/2} A_{i}^{\top} R_{1}^{\top}$, $R_{2} A_{i} W_{i}^{t} A_{i}^{\top} R_{3}^{\top}$, $R_{4} A_{i} (W_{i}^{t})^{1/2}$ and $h_{i}^{t}$ to the server.
\end{itemize}

\paragraph{Global update.} In each global communication round, the detailed process of global update is shown as below:

\begin{itemize}
    \item The server constructs $\wt{P}$ as below 
    {
    \begin{align}\label{eq:alternative_form_of_wt_P}
    \wt{P} = W^{1/2}A^{\top}R_{1}^{\top} R_{1}(R_{2}^{\top}R_{2}AWA^{\top}R_{3}^{\top}R_{3})^{-1}R^{\top}_{4}R_{4}A W^{1/2}
    \end{align}
    }
    \item The server computes $\delta_{x}^{t}$ and $\delta_{s}^{t}$ as below:
    \begin{align*}
        \delta_{x}^{t} = W^{1/2} (I-\wt{P}) W^{1/2} h^{t}, ~~~ \delta_{s}^{t} = \wt{t} \cdot W^{-1/2} \wt{P} W^{1/2} h^{t}
    \end{align*}
    \item The server sends $\delta_{x}^{t}$ and $\delta_{s}^{t}$ to every client.
\end{itemize}

\paragraph{Communication cost.} In this paper, we always assume that $d \geq n$. The Algorithm~\ref{alg:erm_fl} seeds $O(b_{\max} n)$ words at each iteration, where $b_{\max} = \max \{b_{1}, b_{2}, b_{3}, b_{4}\}$. Generally, we choose $b_{\max} = O(\sqrt{n})$. Compared with the naive algorithm mentioned in Model $3$, the Algorithm~\ref{alg:erm_fl} is more practical due to the reason that FL is always limited by the network bandwidth.

\section{Theoretical Analysis} \label{sec:theo}

In our algorithm, the main problem is how to handle the matrix $P$ that is defined as Eq.~\eqref{eq:def_P}, which is used in our IPM where $W \in \R^{n \times n}$ is a block diagonal matrix and $A \in \R^{d \times n}$.

The core of Theorem~\ref{thm:main} is to show the equation $| g^{\top} P h - g^{\top} \wt{P} h |$ could be bounded by $g$, $h$, $A$, and $W$.
In order to prove Theorem~\ref{thm:main}, we divide the proof into the following steps. Given two vectors $g, h \in \R^{d}$. (In the following statement and proof, we assume that $g=h$). We want to prove that
\begin{itemize}
    \item $| g^{\top} P  h - g^{\top} \wh{P} h|$ could be bounded by $g$, $h$, $A$ and $W$. We prove this by using Lemma~\ref{lem:bound_inverse} with $C= W^{1/2} A^\top$, $B=(A W A^\top)^{-1}$, $R= R_1$ and $S = R_2$.
    \item $|g^{\top} \wt{P} h - g^{\top} \wh{P} h|$ could be bounded by $g$, $h$, $A$, $W$, and $\wt{B}$, we prove this by using Lemma~\ref{lem:bound_out_inverse} with $C = W^{1/2}A^{\top}$, $\wt{B}=(R_{1}^{\top}R_{1}AWA^{\top}R_{2}^{\top}R_{2})^{-1}$, $R=R_{3}$, and $S=R_{4}$.
    \item Finally, we could use 
    \begin{align*}
    |g^{\top} P h - g^{\top} \wt{P} h| \leq |g^{\top} P h - g^{\top} \wh{P} h| + |g^{\top} \wh{P} h - g^{\top} \wt{P} h|    
    \end{align*}
    to obtain that $|g^{\top} P h - g^{\top} \wt{P} h|$ is bounded by $g$, $h$, $A$ and $W$.
\end{itemize}

In order to achieve the above-mentioned steps, we need to use the following lemmas. The detailed proof of the following lemma is deferred to Appendix~\ref{sec:bound_error}.

\begin{lemma} \label{lem:second_step}
If the following conditions hold
\begin{itemize}
    \item $\wt{B} \in \R^{d \times d}$ and $C \in \R^{n \times d}$ are two matrices.
    \item $R \in \R^{b_1 \times d}$ and $S \in \R^{b_2 \times d}$ are defined as in Definition~\ref{def:ams_matrix}.
    \item $g \in \R^{n}$ and $h \in \R^{n}$ are vectors.
    \item $b_{\min} = \{b_{1}, b_{2}\}$.
\end{itemize}

Then, we have
\begin{align*}
    g^{\top} C (R^{\top} R) \wt{B} (S^{\top} S) C^{\top} h - g^{\top} C \wt{B} C^{\top} h \lesssim K_0 ,
\end{align*}
with probability at least $1-1/\poly(n)$ and $K_0$ is defined as follows:
\begin{align*}
    K_0 &~:= \frac{\log^{1.5} d}{\sqrt{b_{\min}}} \cdot ( \|g^{\top} C\|_2 \|\wt{B} C^{\top} h\|_2 +  \|g^{\top} C \wt{B} \|_2 \|C^{\top} h\|_2 ) 
    +  \frac{\log^{3} d}{b_{\min}} \cdot \|g^{\top} C \|_2 \|C^{\top} h\|_2 \|\wt{B}\|_F
\end{align*}
\end{lemma}

By using the above lemma, we could obtain the following result by setting $\wt{B}=(R_{1}^{\top}R_{1}AWA^{\top}R_{2}^{\top}R_{2})^{-1}$ and $C=W^{1/2}A^{\top}$, where both $R_{1}$ and $R_{2}$ are independent AMS matrices.
\begin{align*}
 |g^{\top} \wt{P} h - g^{\top} \wh{P} h| \lesssim K_0. 
\end{align*}
Although the above lemma could be used to bound the error of the second step. However, it does not show that $\wt{B}$ could be bounded by $A$ and $W$. It is non-trivial to prove it. 

In order to bound the error, we first use the above lemma to obtain that
\begin{align*}
&~|x^{\top}R^{\top}RB^{-1}S^{\top}Sx - x^{\top}B^{-1}x| \leq \eps_{0} \lambda_{\min}(B^{-1})    
\end{align*}
where $b_{\min} = \{b_{1}, b_{2}\}$, $\kappa = \lambda_{\max}(B) / \lambda_{\min}(B)$, 
$\eps_{0} = O(\sqrt{n} \log^{3} d /b_{\min}) \kappa \in (0,1/10)$, $R \in \R^{b_{1} \times d}$ and $S \in \R^{b_{2} \times d}$ are matrices defined as in Definition~\ref{def:ams_matrix} and $B = (AWA^{\top})^{-1}$.

Then we use the following lemma to bound the error that is caused by adding sketching matrices in the inverse part. 

\begin{lemma} \label{lem:first_step}
If the following conditions hold:
\begin{itemize}
    \item $B \in \R^{d \times d}$ is a matrix.
    \item $R \in \R^{b_{1} \times d}$ and $S \in \R^{b_{2} \times d}$ are defined as in Definition~\ref{def:ams_matrix}.
    \item $g \in \R^{n}$ and $h \in \R^{n}$ are vectors.
    \item $\epsilon_0 \in (0,1/10)$.
\end{itemize}

Then, we have
\begin{align*}
(1-2\eps_{0}) B \preceq (R^{\top}RB^{-1}S^{\top}S)^{-1} \preceq (1+2\eps_{0}) B
\end{align*}
with probability at least $1 - 1/\poly(n)$.
\end{lemma}

By using the above lemma, we could obtain that
\begin{align*}
&~ | g^{\top} P  h - g^{\top} \wh{P} h| \leq 2 \eps_{0} \|g^{\top} C\|_{2} \|C^{\top} h\|_{2} \|B\|_{2}
\end{align*}
with probability $1-1/\poly(n)$, where $C=W^{1/2}A^{\top}$ and $B=(AWA^{\top})^{-1} \in \R^{d \times d}$.

Finally, we combine the result of Lemma~\ref{lem:second_step} and Lemma~\ref{lem:first_step} together to get the following theorem.

\begin{theorem}\label{thm:main}
If the following conditions hold
\begin{itemize}
    \item Given $A \in \R^{d \times n}$ and $W \in \R^{n \times n}$.
    \item Let $R_{1} \in \R^{b_{1} \times d}$, $R_{2} \in \R^{b_{2} \times d}$, $R_{3} \in \R^{b_{3} \times d}$ and $R_{4} \in \R^{b_{4} \times d}$ be four independent AMS matrices.
    \item $P$ is defined as in Eq.~\eqref{eq:def_P}.  
    \item $\wt{P}$ is defined as in Def.~\ref{def:app_P}.
    \item Let $g,h \in \R^{n}$ be two vectors.
    \item Let $b_{\min} : = \min \{b_1,b_2\}$. 
\end{itemize}

Then, we have  
\begin{align*}
  |g^{\top} P h - g^{\top} \wt{P} h| 
  \lesssim \log^6 d \cdot  (\frac{1}{\sqrt{b_{\min}}}  + \frac{n }{b_{\min}^{2}})  
   \cdot \kappa \cdot \|g^{\top}C\|_{2} \|C^{\top} h\|_{2} \|B\|_{2}  
\end{align*}
with probability at least $1-1/\poly(n)$.  Note that $C = W^{1/2}A^{\top}$, $B=(AWA^{\top})^{-1}$, and $\kappa = \lambda_{\max}(B) / \lambda_{\min}(B)$.
\end{theorem}

Due to the reason that $\kappa > 1$, we could choose that $b_{\min} = \eps^{-1} \sqrt{n} \kappa^{2} \log^{3} d$ and $\eps \in (0,1/10)$. Then, we could obtain that
$
 |g^{\top} P h - g^{\top} \wt{P} h| \leq \eps \|g^{\top}C\|_{2} \|C^{\top} h\|_{2} \|B\|_{2}.
$

\section{Compared to Standard Methods} \label{sec:comparison}

In order to show the effectiveness and efficiency of our algorithm, we discuss the following three naive models and point out the disadvantages of each model respectively.

We introduce the following three straightforward methods: Model 1 and Model 2 cannot get the right result under their respective framework. Model 3 can get the correct result, but it needs to send $O(n^{2})$ words at each iteration. Moreover, Model 3 also requires clients to share their data with the untrusted server, which is not allowed under FL setting.

\textbf{Model 1:} In the $t$-th step, each client does the following operations: (1) Compute $W_{i}^{t}$ and $h_{i}$; (2) Compute local $P_{i}^{t}$, where $P_{i}^{t} = (W_{i}^{t})^{1/2} A_{i}^{\top} (A_{i} W_{i}^{t} A_{i}^{\top})^{-1} A_{i} (W_{i}^{t})^{1/2}$; (3) By using local $P_{i}^{t}$ and $h_{i}$, the client could compute local update $\delta_{x,i}$ and $\delta_{s,i}$; (4) Finally, client sends its local update $\delta_{x,i}$ and $\delta_{s,i}$ to the server.

The Server combines all gradients together. However, the main issue is that
\begin{align*}
\oplus_{i=1}^{m}[(W_{i}^{t})^{1/2}(I-P_{i}^{t})(W_{i}^{t})^{1/2} h_{i}] \neq (W^{t})^{1/2}(I-P^{t})(W^{t})^{1/2}h
\end{align*}
and
\begin{align*}
\oplus_{i=1}^{m} [(W_{i}^{t})^{-1/2}P_{i}^{t}(W_{i}^{t})^{1/2} h_{i}] \neq (W^{t})^{-1/2}P^{t}(W^{t})^{1/2}h
\end{align*}
where $P=(W^{t})^{1/2} A^{\top} (A W^{t} A^{\top})^{-1} A (W^{t})^{1/2}$, and $I$ is an identify matrix.

\textbf{Model 2:} In the $t$-th step, each client does the following operations: (1) Compute $W_{i}^{t}$ and $h_{i}$ locally; (2) Send $(A_{i}W_{i}^{t}A_{i}^{\top})^{-1}$ and $h_{i}$ to the server.
However, this method does not work well. The reason is that 
\begin{align*}
&~\otimes_{i=1}^{m} [(W_{i}^{t})^{1/2} A_{i}^{\top} (A_{i}W_{i}^{t}A_{i}^{\top})^{-1} A_{i} (W_{i}^{t})^{1/2}] \oplus_{i=1}^{m} h_{i} \\
&~\neq [(W^{t})^{1/2} A^{\top} (AW^{t}A^{\top})^{-1} A (W^{t})^{1/2}] h
\end{align*}
where 
$
W^{t} = \otimes_{i=1}^{m} W_{i}^{t}, A = \oplus_{i=1}^{m} A_{i}, \mathrm{~~~and~~~} h = \oplus_{i=1}^{m} h_{i}.
$

\textbf{Model 3:} Each client sends data to the server at the $0$-th iteration. Then, in the $t$-th step, each client does the following operations: (1) Compute locally $W_{i}^{t}$ and $h_{i}$; (2) Send $W_{i}^{t}$ and $h_{i}$ to the server.

The server computes $P$ by the following equation:

\begin{align*}
P=(\otimes_{i=1}^{m} (W_{i}^{t})^{1/2} A^{\top}) (A \otimes_{i=1}^{m} W_{i}^{t} A^{\top})^{-1} (A \otimes_{i=1}^{m} (W_{i}^{t})^{1/2})
\end{align*}

Compared to the above-mentioned two models, this method could get the correct result in the end. However, it has to send $O(n^2)$ words at each iteration. In reality, the distributed machine learning is always limited by the network bandwidth. Moreover, people usually are not willing to share their private data with the untrusted system because of data privacy. In view of this, we propose a communication-efficient distributed interior point method under FL.

\section{Conclusion and Discussion} \label{sec:discussion}

In a nutshell, we present the first distributed interior point method algorithm (\textsc{FL-IPM}) that is used to address empirical risk minimization under FL. There are differences between our algorithm and existing algorithms and the novelty of our work is shown below: (1) There exist a large number of works related to the distributed first-order optimization algorithms. However, our algorithm is a second-order optimization problem under federated learning settings. (2) We use the sketching technique to reduce the communication cost of federated learning, which is the bottleneck of federated learning. (3) Compared with the existing distributed second-order optimization algorithms, we can provide convergence analysis for our solution without making strong assumptions.

As for future work, there are several things we need to consider, if we want to apply our algorithm in the real system: First, we need to consider the stragglers and device heterogeneity in the real system environment. We need to design robust algorithms to deal with stragglers during the training. In addition, the scalability of large networks is also very important, especially the latency and throughput of the network. Finally, the computational cost of the devices and server should be taken into consideration. We present theoretical results in this paper, and we are not aware of any negative societal impact.

\ifdefined\isarxiv

\else
\bibliography{ref}
\bibliographystyle{plain}

\fi

\newpage
\appendix
\onecolumn
\section*{Appendix}

\paragraph{Roadmap.} 
The structure of the appendix is outlined as follows:
\begin{itemize}
    \item Section~\ref{sec:fastjl} claims the probability tools used in this paper and shows the properties of random sketching matrix.
    \item Section~\ref{sec:sketch_more} presents how to bound the error of adding two sketching matrices.
    \item Section~\ref{sec:bound_error} shows that $|g^{\top} P h - g^{\top} \wt{P} h|$ is small. 
    \item Section~\ref{sec:proof_main_res} presents the primary outcome of this paper along with its corresponding proof.
    \item Section~\ref{sec:central_path} shows several basic results of Algorithm~\ref{alg:erm_fl}.
    \item Section~\ref{sec:initial_point_and_termination_condi} states some basic results of self-concordance function.
\end{itemize}

\section{Probability Tools and Basic Properties of Random Sketching Matrices}\label{sec:fastjl}

In this paper, we care less about the running time of each client in our application. The issue we want to address in this paper is the limitation of the network bandwidth (bandwidth between server and clients). In view of this, we use subsampled randomized Hadamard/Fourier matrix\footnote{We want to remark that SRHT has fast computation advantage compared to AMS. Using SRHT~\cite{ldfu13} allows multiplying the matrix with $k$ vectors only takes $k n \log n$ time. This is much faster compared to AMS. In our application, we only use nice statistical properties of SRHT matrices without  using any fast Fourier transform~\cite{ct65}, or more fancy sparse Fourier transform~\cite{hikp12b,hikp12a,price13,ikp14,ik14,ps15,ckps16,k16,k17,nsw19,gss22,sswz22_quartic,sswz22_lattice}.} and AMS matrices.

The basic ideas of handling randomness in sketching matrices have been used in a number of previous work \cite{psw17,lsz19,jswz21,sy21,syyz23}. However, in our case, we have more different sketching matrices and also need to apply sketching matrices inside inversion.

\begin{figure}[!ht]
    \centering
    \includegraphics[width=0.6\linewidth]{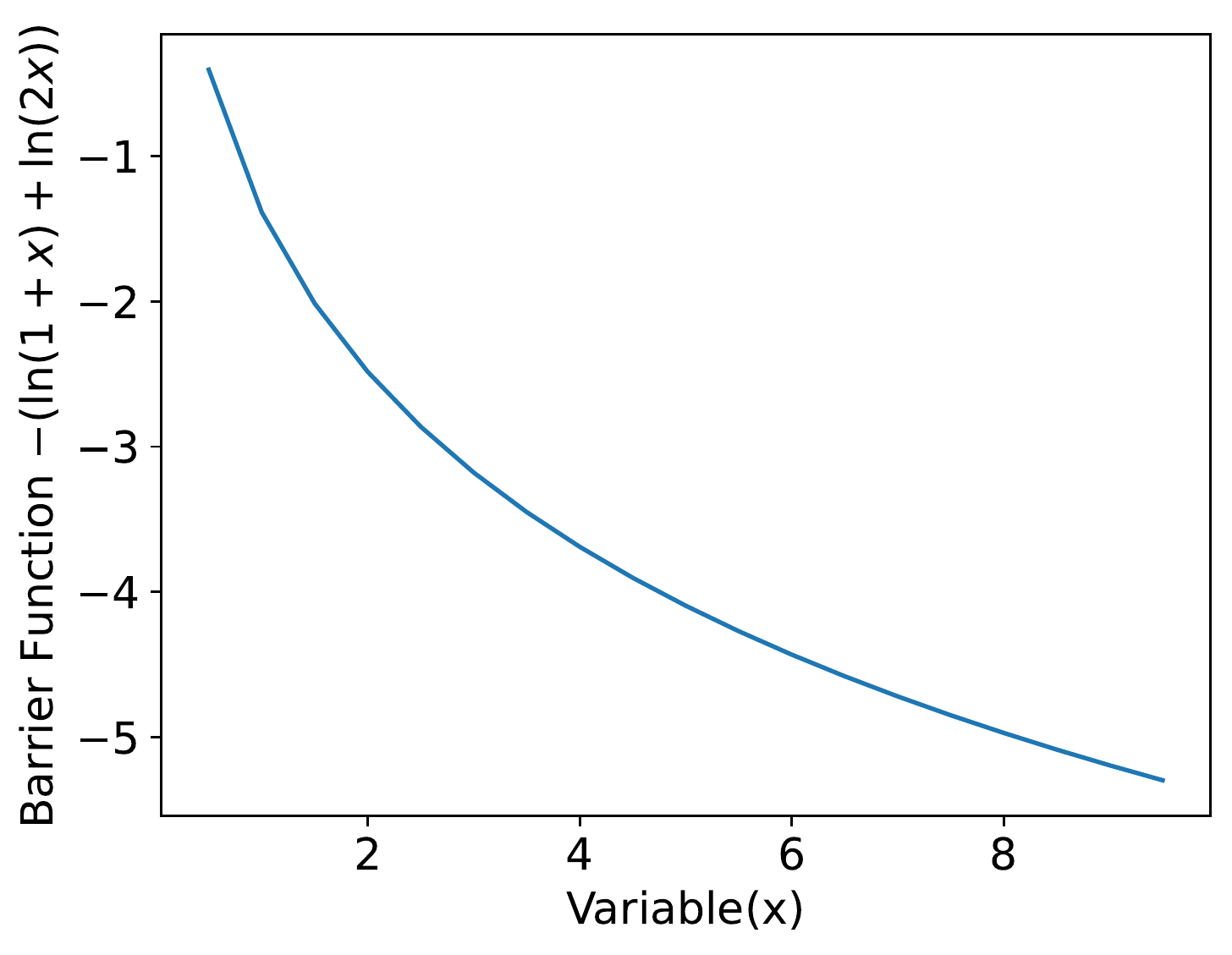}
    \caption{An example of the barrier function: $-(\ln(1+x) + \ln(2x))$. The variable is changed from $0.5$ to $10$. In this case, $A = [-1, -2]^{\top}$, and $b=[1, 0]^{\top}$.} 
    \label{fig:barrier_function}
\end{figure}

In Section~\ref{sub:fastjl:inequalities}, we introduce the concentration inequalities. In Section~\ref{sub:fastjl:properties}, we present the properties obtained from random projection.

\subsection{Concentration inequalities}
\label{sub:fastjl:inequalities}

We first state several useful inequalities.

\begin{lemma}[Lemma 1 on page 1325 of \cite{lm00}]\label{lem:chi_squared}
If the following conditions hold
\begin{itemize}
    \item $X \sim {\cal X}_k^2$ is a random variable, which is a chi-squared distribution and has $k$ degrees of freedom.
    \item Each of them has a mean of $0$ and a variance of $\sigma^2$.
\end{itemize}

Then, we have
\begin{align*}
\Pr[ X - k \sigma^2 \geq (2 \sqrt{ k t } + 2 t ) \sigma^2 ] \leq \exp( - t )
\end{align*}
and
\begin{align*}
\Pr[ k \sigma^2 - X \geq 2 \sqrt{ k t } \sigma^2 ] \leq \exp( - t ).
\end{align*}
\end{lemma}

\begin{lemma}[Khintchine's Inequality]
\label{lem:khintchine_inequality}
If the following conditions hold
\begin{itemize}
    \item $\sigma_1, \cdots, \sigma_n$ are the independent and identically distributed sign random variables.
    \item $z_1, \cdots, z_n$ are real numbers.
\end{itemize}

Then, there exists positive constants, namely $C$ and $C'$, satisfying that:
\begin{align*}
\Pr \left[ \left| \sum_{i=1}^n z_i \sigma_i \right| \geq C t \| z \|_2 \right] \leq \exp( - C' t^2 )
\end{align*}
\end{lemma}

\begin{lemma}[Bernstein Inequality]\label{lem:bernstein_inequality}

If the following conditions hold
\begin{itemize}
    \item $X_1, \cdots, X_n$ is a set of independent random variables with zero means. 
    \item For any arbitrary $1 \leq i \leq n$, let the absolute value of each $X_i$ is almost surely bounded by a constant $M$. 
\end{itemize}

Then, for any positive value $t$, the following inequality holds:
\begin{align*}
\Pr \left[ \sum_{i=1}^n X_i > t \right] \leq \exp \left( - \frac{ t^2 / 2 }{ \sum_{j=1}^n \E [X_j^2] + M t / 3 } \right)
\end{align*}
\end{lemma}

\subsection{Properties obtained by random projection}
\label{sub:fastjl:properties}

Here, we formally define the SRHT matrix and AMS sketching matrix and analyze their properties.

\begin{definition}[Subsampled randomized Hadamard/Fourier transform (SRHT) matrix~\cite{ldfu13}]
\label{def:srht}
The SRHT matrix, denoted as $R = \sqrt{n/b} \cdot SHD$, where $R \in \mathbb{R}^{b \times n}$, and $S \in \mathbb{R}^{b \times n}$ represents a random matrix whose rows are $b$ uniform samples (without replacement) from the standard basis of $\R^{n}$, $H \in \R^{n \times n}$ is a normalized Walsh-Hadamard matrix, and $D \in \R^{n \times n}$ is a diagonal matrix whose diagonal elements are i.i.d. Rademacher random variables.
\end{definition}

\begin{definition}[AMS sketch matrix~\cite{ams99}]
\label{def:ams}
Let $h_{1}, h_{2}, \dots, h_{b}$ be $b$ random hash functions picking from a $4$-wise independent hash family $\mathcal{H}=\{h:[n] \rightarrow \{-1/\sqrt{b}, +1/\sqrt{b}\}\}$. Then, $R \in \R^{b \times n}$ is a AMS sketch matrix if we set $R_{i,j}=h_{i}(j)$.
\end{definition}

\begin{lemma}[Lemma E.5 in~\cite{lsz19}]
\label{lem:sketch_vector}
If the following conditions hold
\begin{itemize}
    \item Let $h \in \R^n$ be a fixed vector.
    \item Let $R \in \R^{b \times n}$ be a SRHT or AMS sketch matrix as in Definition~\ref{def:srht} and~\ref{def:ams}.
\end{itemize}
Then, we have
\begin{align*}
& ~ \E[ R^\top R h ] = h, ~~~ 
\E[ (R^\top R h)_i^2 ] \leq h_i^2 + \frac{1}{b} \| h \|_2^2 \\
& ~ \Pr \left[ | ( R^\top R h )_i - h_i | > \| h \|_2 \frac{\log ( n / \delta ) }{ \sqrt{b} } \right] \leq \delta
\end{align*}
\end{lemma}

\section{Sketch more than once}
\label{sec:sketch_more}

\begin{figure}[!ht]
    \centering
    \includegraphics[width=0.5\textwidth]{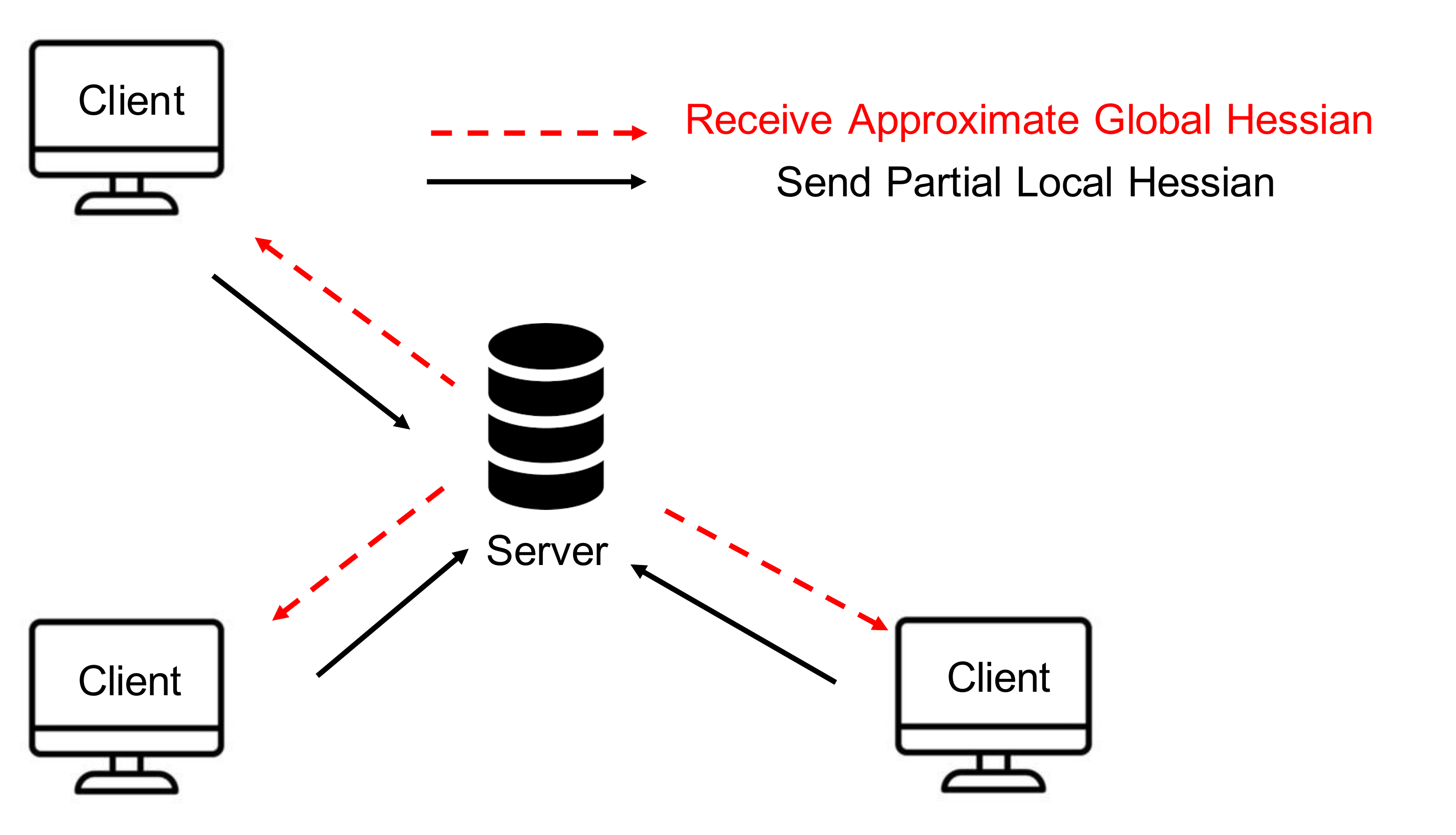}
    \caption{This is an overview of our framework. In our framework, there is no need for clients to share data with the server. Clients share partial Hessian information with the server. And the server computes the update information by using Hessian information, then sends the update information to the client.}
    \label{fig:overview}
\end{figure}

Now, we can bound the error of adding two sketching matrices.

\begin{lemma}[Error bound of adding two sketching matrices]
\label{lem:CE_two_sketch}
If the following conditions hold
\begin{itemize}
    \item $R \in \R^{b_1 \times n}, S \in \R^{b_2 \times n}$ are defined as in Def.~\ref{def:ams}. 
    \item $B \in \R^{n \times n}$ is a matrix.
    \item $u,v \in \R^n$ are vectors.
\end{itemize}

Then, with probability $1-1/\poly(n)$,
\begin{align*}
  &~| u^{\top} R^\top R B S^{\top} S v - u^{\top} B v | \\
\lesssim &~ \frac{\log^{1.5} n}{\sqrt{b_1}} \cdot \|u\|_2 \|B v\|_2 
+ \frac{\log^{1.5} n}{\sqrt{b_2}} \cdot \|u^{\top} B\|_2 \|v\|_2 
+ \frac{\log^{3} n}{\sqrt{b_1 b_2}} \cdot \|u\|_2 \|v\|_2 \|B\|_F.
\end{align*}
holds.
\end{lemma}

\begin{proof}
Let $i$ be in $[n]$. 

Let the $i$-th column of $R$ be $R_i \in \R^{b_1}$.

Let the $i$-th column of $S$ be $S_i \in \R^{b_2}$.

Let $\sigma_i$ be a random sign.

Let $R$ be an AMS matrix.

Every column $R_i$ of $R$ follows the same distribution as $\sigma_i R_i$.

We have that $R$ satisfies:
\begin{align}
    1. &~ \langle R_i, R_i\rangle = 1, \forall i \in [n]. \label{eq:two_sketch_same_Ri} \\
    2. &~ \Pr [\langle R_i, R_j\rangle \leq \frac{\sqrt{\log(n/\delta)}}{\sqrt{b_1}}, \forall i\neq j \in [n] ] \geq 1-\delta. \label{eq:two_sketch_diff_Ri}
\end{align}

Likewise, $S$ is an AMS matrix, and the distribution of each column $S_i$ of $S$ is identical to $\sigma'_i S_i$, where $\sigma'_i$ represents a random sign. Additional information can be found in \cite{ams99}.

Then, we can get
\begin{align}\label{eq:two_sketch_close_form_whp}
    ~u^{\top} (R^{\top} R) B (S^{\top} S) v 
    =~ \sum_{i,j,i',j'} u_{i} v_{j'} \sigma_{i} \sigma_{j} \sigma'_{i'} \sigma'_{j'} \langle R_i, R_j \rangle B_{j,i'} \langle S_{i'}, S_{j'} \rangle
\end{align}

Therefore, we can divide the summation in equation Eq.~\eqref{eq:two_sketch_close_form_whp} into three components:
\begin{enumerate}
    \item The first part involves two pairs of indices being identical: $j = i$ and $j' = i'$.
    \item The second part occurs when one pair of indices is the same: either $j = i$ and $j' \neq i'$, or conversely, $j \neq i$ and $j' = i'$.
    \item The third part arises when no pair of indices are the same: $j \neq i$ and $j' \neq i'$.
\end{enumerate}

{\bf Proof of Part 1.} 

Suppose $j = i$ and $j' = i'$. 

We can get
\begin{align}\label{eq:two_sketch_whp_part_1}
    &~\sum_{i=j,i'=j'} u_{i} v_{j'} \sigma_{i} \sigma_{j} \sigma'_{i'} \sigma'_{j'} \langle R_i, R_j \rangle B_{j,i'} \langle S_{i'}, S_{j'} \rangle \notag \\
    = &~ \sum_{i, i'} u_{i} v_{i'} B_{i,i'} \notag \\ 
    = &~ u^{\top} B v
\end{align}
For the first step, we use the fact that $\langle R_i, R_i \rangle = \langle S_{i'}, S_{i'} \rangle = 1$ for all $i$ and $i'$ in $[n]$, as shown in Eq.~\eqref{eq:two_sketch_same_Ri}.

{\bf Proof of Part 2.} 

Suppose that either $j = i$ and $j' \neq i'$, or conversely, $j \neq i$ and $j' = i'$.

Without loss of generality, we suppose $j = i$ and $j' \neq i'$. 

Then, we can get
\begin{align*}
    &~ \sum_{i=j, i'\neq j'} u_{i} v_{j'} \sigma_{i} \sigma_{j} \sigma'_{i'} \sigma'_{j'} \langle R_i, R_j \rangle B_{j,i'} \langle S_{i'}, S_{j'} \rangle \\
    = &~ \sum_{i, i'\neq j'} u_{i} v_{j'} \sigma'_{i'} \sigma'_{j'} B_{i,i'} \langle S_{i'}, S_{j'} \rangle  \\
    = &~ \sum_{j'} \sigma'_{j'} v_{j'} \sum_{i'\neq j'} \sigma'_{i'} (B^{\top} u)_{i'} \langle S_{i'}, S_{j'}\rangle,
\end{align*}
For the first step, we use the fact that $\langle R_i, R_i \rangle = 1$ for all $i$ in $[n]$, as shown in Eq.~\eqref{eq:two_sketch_same_Ri}. For the second step, we use $\sum_i u_i B_{i,i'} = (B^{\top} u)_{i'}$.

By the Union bound and Lemma~\ref{lem:khintchine_inequality}, we can get
\begin{align*}
    &~ (\sum_{j'} \sigma'_{j'} v_{j'} \sum_{i'\neq j'} \sigma'_{i'} (B^{\top} u)_{i'} \langle S_{i'}, S_{j'}\rangle )^2 \\
    \lesssim &~ \log n \cdot \sum_{j'} v_{j'}^2 (\sum_{i'\neq j'} \sigma'_{i'} (B^{\top} u)_{i'} \langle S_{i'}, S_{j'} \rangle )^2 \\
    \lesssim &~ \log^2 n \cdot \sum_{j'} v_{j'}^2 \sum_{i' \neq j'} (B^{\top} u)_{i'}^2 \langle S_{i'}, S_{j'}\rangle^2 \\
    \lesssim &~ \log^3 n / b_2 \cdot \sum_{j'} v_{j'}^2 \sum_{i' \neq j'} (B^{\top} u)_{i'}^2 \\
    \lesssim &~ \log^3 n / b_2 \cdot \|v\|_2^2 \|B^{\top} u\|_2^2,
\end{align*}
with probability at least $1 - 1/\poly(n)$, where the first step follows from $t=O(\sqrt{\log n})$ and Lemma~\ref{lem:khintchine_inequality}, the second step is obtained by $t=O(\sqrt{\log n})$ and Lemma~\ref{lem:khintchine_inequality} again, and the third step is derived from Eq.~\eqref{eq:two_sketch_diff_Ri}.

Combining the previous two equations, and considering the symmetry of the case where $i'=j'$ and $i\neq j$, we can get that
\begin{align}\label{eq:two_sketch_whp_part_2}
    &~ \sum_{\substack{i=j, i'\neq j' \\\mathrm{or~} i'=j', i\neq j}} u_{i} v_{j'} \sigma_{i} \sigma_{j} \sigma'_{i'} \sigma'_{j'} \langle R_i, R_j \rangle B_{j,i'} \langle S_{i'}, S_{j'} \rangle \notag \\
    \lesssim &~ \log^{1.5} n / \sqrt{b_1} \cdot \|u\|_2 \|B v\|_2 
    + \log^{1.5} n / \sqrt{b_2} \cdot \|u^{\top} B\|_2 \|v\|_2
\end{align}
with a probability of at least $1 - 1/\poly(n)$.

{\bf Proof of Part 3.} 

Suppose $j \neq i$ and $j' \neq i'$. 

We can show
\begin{align*}
    &~ ( \sum_{i \neq j, i' \neq j'} u_{i} v_{j'} \sigma_{i} \sigma_{j} \sigma'_{i'} \sigma'_{j'} \langle R_i, R_j \rangle B_{j,i'} \langle S_{i'}, S_{j'} \rangle )^2 \\
    = &~ ( \sum_{i} \sigma_i u_i \sum_{j'} \sigma'_{j'} v_{j'} \sum_{i' \neq j'} \sigma'_{i'} \langle S_{i'}, S_{j'} \rangle \sum_{j \neq i} \sigma_i \langle R_i, R_j \rangle B_{j,i'} )^2 \\
    \lesssim &~ \log n \cdot \sum_{i} u_i^2 ( \sum_{j'} \sigma'_{j'} v_{j'} \sum_{i' \neq j'} \sigma'_{i'} \langle S_{i'}, S_{j'} \rangle \sum_{j \neq i} \sigma_i \langle R_i, R_j \rangle B_{j,i'} )^2 \\
    \lesssim &~ \log^2 n \cdot \sum_{i} u_i^2 \sum_{j'} v_{j'}^2 ( \sum_{i' \neq j'} \sigma'_{i'} \langle S_{i'}, S_{j'} \rangle \sum_{j \neq i} \sigma_i \langle R_i, R_j \rangle B_{j,i'} )^2 \\
    \lesssim &~ \log^3 n \cdot \sum_{i} u_i^2 \sum_{j'} v_{j'}^2 \sum_{i' \neq j'} \langle S_{i'}, S_{j'} \rangle^2 ( \sum_{j \neq i} \sigma_i \langle R_i, R_j \rangle B_{j,i'} )^2 \\
    \lesssim &~ \log^4 n \cdot \sum_{i} u_i^2 \sum_{j'} v_{j'}^2 \sum_{i' \neq j'} \langle S_{i'}, S_{j'} \rangle^2 \sum_{j \neq i} \langle R_i, R_j \rangle^2 B_{j,i'}^2 \\
    \lesssim &~ \log^6 n / (b_1 b_2) \cdot \|u\|_2^2 \|v\|_2^2 \|B\|_F^2,
\end{align*}
with probability $1 - 1/\poly(n)$, where 2nd step follows from $t = O(\sqrt{n})$ and Lemma~\ref{lem:khintchine_inequality}, the 3rd comes from $t = O(\sqrt{n})$, Lemma~\ref{lem:khintchine_inequality}, for all $i \in [n]$, and employing the Union bound to combine the $n$ inequalities, the 4th and 5th step can be justified based on the same reasoning as the 3rd step. 

For the 6th step, we use the fact that for all $i' \neq j' \in [n]$ and $i\neq j \in [n]$, with a probability of at least $1-1/\poly(n)$, 
\begin{align*}
    \langle S_{i'}, S_{j'} \rangle \lesssim \sqrt{(\log n) / b_2}
\end{align*}
and 
\begin{align*}
    \langle R_i, R_j\rangle \lesssim \sqrt{(\log n) / b_1}.
\end{align*}
For all $i$, $j$, $i'$, and $j'$ in $[n]$, we apply the Union bound to combine $2n^2$ such bounds.

Therefore, we can get
\begin{align}\label{eq:two_sketch_whp_part_3}
    ~ \sum_{i \neq j, i' \neq j'} u_{i} v_{j'} \sigma_{i} \sigma_{j} \sigma'_{i'} \sigma'_{j'} \langle R_i, R_j \rangle B_{j,i'} \langle S_{i'}, S_{j'} \rangle 
    \lesssim ~ \log^3 n / \sqrt{b_1 b_2} \cdot \|u\|_2 \|v\|_2 \|B\|_F.
\end{align}
with probability at least $1 - 1/\poly(n)$.

{\bf Combining Part 1, Part 2, and Part 3.} 

First, we add Eq.~\eqref{eq:two_sketch_whp_part_1}, \eqref{eq:two_sketch_whp_part_2}, and \eqref{eq:two_sketch_whp_part_3} together. 

Then, we plug their sum into Eq.~\eqref{eq:two_sketch_close_form_whp}. 

Finally, through Union bound, we can get
\begin{align*}
&~ u^{\top} (R^{\top} R) B (S^{\top} S) v - u^{\top} B v \\
\lesssim &~ \frac{\log^{1.5} n}{\sqrt{b_1}} \cdot \|u\|_2 \|B v\|_2 
+ ~ \frac{\log^{1.5} n}{\sqrt{b_2}} \cdot \|u^{\top} B\|_2 \|v\|_2 
+ ~ \frac{\log^{3} n}{\sqrt{b_1 b_2}} \cdot \|u\|_2 \|v\|_2 \|B\|_F,
\end{align*}
with probability at least $1 - 1/\poly(n)$.

Therefore, we complete the proof.
\end{proof}
\section{Bounding error of sketching}
\label{sec:bound_error}

This section is arranged as follows:

\begin{itemize}
    \item Section~\ref{sec:def_P} gives the definition of $P$, $\wh{P}$, and $\wt{P}$.
    \item Section~\ref{sec:part_one} presents the steps to prove that $P \approx \wt{P}$.
    \item Section~\ref{sec:bound_P_wh_P} shows that $|g^{\top} P h - g^{\top} \wh{P} h|$ is small.
    \item Section~\ref{sec:tool_bound_P_wh_P} presents the tools that we use to bound $|g^{\top} P h - g^{\top} \wh{P} h|$.
    \item Section~\ref{sec:bound_wt_P_wh_P} shows that $|g^{\top} \wt{P} h - g^{\top} \wh{P} h|$ is small.
    \item Section~\ref{sec:tool_bound_wt_P_wh_P} presents the tools that we use to bound $|g^{\top} \wt{P} h - g^{\top} \wh{P} h|$.
    \item Section~\ref{sec:bound_P_wt_P} shows that $| g^{\top} P h - g^{\top} \wt{P} h|$ is small by combining the result of $|g^{\top} P h - g^{\top} \wh{P} h|$ and $|g^{\top} \wt{P} h - g^{\top} \wh{P} h|$.
\end{itemize}

\subsection{Definition of \texorpdfstring{$P$}{}, \texorpdfstring{$\wh{P}$}{}, and \texorpdfstring{$\wt{P}$}{}} \label{sec:def_P}

In this section, we formally define $P$, $\wh{P}$, and $\wt{P}$.

\begin{definition}[Definition of Projection Matrices]
We define $P \in \R^{n \times n}$, $\wh{P} \in \R^{n \times n}$, and $\wt{P} \in \R^{n \times n}$ as follows:
\begin{align*}
    P := & ~ W^{1/2}A^{\top}(AWA^{\top})^{-1}AW^{1/2} \\
    \wh{P} := & ~ W^{1/2}A^{\top}(R_1^\top R_1 AWA^{\top} R_2^\top R_2 )^{-1}AW^{1/2}\\
    \wt{P} := & ~ W^{1/2} A^\top R_3^\top R_3 ( R_1^\top R_1 A W A^\top R_2^\top R_2 )^{-1} R_4^\top R_4 A W^{1/2}
\end{align*}
where $R_{1} \in \R^{b_{1} \times d}$, $R_{2} \in \R^{b_{2} \times d}$, $R_{3} \in \R^{b_{3} \times d}$, and $R_{4} \in \R^{b_{4} \times d}$ are sketching matrices.
\end{definition}
Among them, $P$ is the ideal case of the projection matrix. $\wt{P}$ is the projection matrix we use under FL. We construct $\wh{P}$ to analyze that $|g^{\top} P h - g^{\top} \wt{P} h|$ is bounded by $g$, $h$, $A$, and $W$, for any $g \in \R^{n}$ and $h \in \R^{n}$. 

\subsection{Proof sketch}\label{sec:part_one}

In this section, we show that $P \approx \wt{P}$.

Our goal is to show that
\begin{align*}
    |g^{\top} P h - g^{\top} \wt{P} h|
\end{align*}
is bounded by $g$, $h$, $A$ and $W$.

We split it into following steps. For any two vectors $g, h \in \R^{d}$, we want to prove that
\begin{itemize}
    \item $| g^{\top} P  h - g^{\top} \wh{P} h|$ is small, we prove this by using Lemma~\ref{lem:bound_inverse} with 
    \begin{itemize}
        \item $C= W^{1/2} A^\top$,
        \item $B=(A W A^\top)^{-1}$, 
        \item $R= R_1$, and 
        \item $S = R_2$.
    \end{itemize}
    \item $|g^{\top} \wt{P} h - g^{\top} \wh{P} h|$ is small, we prove this by using Lemma~\ref{lem:bound_out_inverse} with 
    \begin{itemize}
        \item $B=(R_{1}^{\top}R_{1}AWA^{\top}R_{2}^{\top}R_{2})^{-1}$, 
        \item $C = W^{1/2}A^{\top}$, 
        \item $R=R_{3}$, and 
        \item $S=R_{4}$.
    \end{itemize}
    \item $|g^{\top} P h - g^{\top} \wt{P} h|$ is small, we could prove it by using 
    \begin{itemize}
        \item $|g^{\top} P h - g^{\top} \wt{P} h| \leq |g^{\top} P h - g^{\top} \wh{P} h| + |g^{\top} \wh{P} h - g^{\top} \wt{P} h|$, 
        \item $\|\wt{B}C^{\top}h\|_{2} \leq \|\wt{B}\|_{2} \|C^{\top} h\|_{2} \leq (1+\eps_{0})\|B\|_{2} \|C^{\top} h\|_{2}$, and
        \item $\|\wt{B}\|_{F} \leq \sqrt{n} \|\wt{B}\|_{2} \leq (1+\eps_{0}) \sqrt{n} \|B\|_{2}$
    \end{itemize}
\end{itemize}

\subsection{Bounding \texorpdfstring{$|g^{\top} P  h - g^{\top} \wh{P} h|$}{}} \label{sec:bound_P_wh_P}

The goal of this section is to prove the following lemma to indicate that we could bound $|g^{\top} P h - g^{\top} \wh{P} h|$. Note that we assume that $g=h$ in this lemma. However, in order to make other lemma more general, we do not assume that $g=h$ in other lemma in this section.
\begin{lemma}[$P$ and $\wh{P}$ are close]
If the following conditions hold
\begin{itemize}
    \item Let $g \in \R^{n}$ and $h \in \R^{n}$ be two vectors. 
    \item Let $\epsilon_0 \in (0,1/10)$.
\end{itemize}

Then, we have
\begin{align*}
| g^\top P h - g^\top \wh{P} h | \leq &~ 2 \eps_{0} g^{\top} C B C^{\top} h \\
\leq &~ 2 \eps_{0} \|g^{\top} C \|_{2} \|C^{\top} h\|_{2} \|B\|_{2}
\end{align*}
with probability at least $1- 1/\poly(n)$, where $C=W^{1/2} A^{\top}$ and $B = (AWA^{\top})^{-1} \in \R^{d \times d}$. 
\end{lemma}
\begin{proof}
We assume that $f(B, R, S) = R^{\top}RB^{-1}S^{\top}S$. By using Lemma~\ref{lem:bound_inverse}, we could obtain that
\begin{align*}
&~ (1-2\eps_{0}) B \preceq (f(B,R,S))^{-1} \preceq (1+2\eps_{0}) B
\end{align*}
Then, for any two vectors $g, h \in \R^{d}$, we could obtain that 
\begin{align*}
(1-2\eps_{0}) g^{\top} C B C^{\top} h \leq &~ g^{\top} C (f(B,R,S))^{-1} C^{\top} h \\
\leq &~ (1+2\eps_{0}) g^{\top} C B C^{\top} h
\end{align*}
According to the above inequality, it is easy for us to get that $|g^\top P h - g^\top \wh{P} h| \leq 2 \eps_{0} g^{\top} C B C^{\top} h$. 

We could obtain that 
\begin{align*}
g^{\top} C B C^{\top} h \leq &~ \sqrt{g^{\top} C B C^{\top} g} \cdot \sqrt{ h^{\top} C B C^{\top} h} \\
= &~ \|g^{\top} C\|_{2} \|C^{\top} h\|_{2} \|B\|_{2} 
\end{align*}
by using the Cauchy-Schwarz inequality. Therefore, we could get that 
\begin{align*}
&~ |g^{\top} P h - g^{\top} \wh{P} h| \leq 2 \eps_{0} \|g^{\top} C\|_{2} \|C^{\top} h\|_{2} \|B\|_{2}
\end{align*}
This finishes the proof.
\end{proof}

\subsection{Tools for bounding \texorpdfstring{$| g^{\top} P  h - g^{\top} \wh{P} h|$}{}} \label{sec:tool_bound_P_wh_P}

In this section, we present the tools for bounding $| g^{\top} P  h - g^{\top} \wh{P} h|$.

\begin{lemma}[Tools for showing $P$ and $\wh{P}$ are close] \label{lem:bound_inverse}
If the following conditions hold
\begin{itemize}
    \item $R \in \R^{b_{1} \times d}$, $S \in \R^{b_{2} \times d}$ are defined as in Definition~\ref{def:ams_matrix}.
    \item $g,h \in \R^{n}$ are two vectors.
\end{itemize}

Then, we have that 
\begin{align*}
(1-2\eps_{0}) B \preceq (R^{\top}RB^{-1}S^{\top}S)^{-1} \preceq (1+2\eps_{0}) B
\end{align*}
with probability at least $1 - 1/\poly(n)$, where $\eps_{0} \in (0, 1/10)$, and $B \in \R^{d \times d}$.
\end{lemma}

\begin{proof}

Given any $x \in \R^{d}$ such that $\|x\|_{2} = 1$, we could use Lemma~\ref{lem:CE_two_sketch} to prove that 
\begin{align*}
    |x^{\top}R^{\top}RB^{-1}S^{\top}Sx - x^{\top}B^{-1}x| \leq \eps_{0} \lambda_{\min}(B^{-1}),
\end{align*}

where $b_{\min} = \{b_{1}, b_{2}\}$, $\kappa = \lambda_{\max}(B) / \lambda_{\min}(B)$ and $\eps_{0} = O(\sqrt{n} \log^{3} d /b_{\min}) \kappa$. Then, we have to prove two cases:

\textbf{Case 1:} From $|x^{\top}R^{\top}RB^{-1}S^{\top}Sx - x^{\top}B^{-1}x| \leq \eps_{0} \lambda_{\min}(B^{-1})$, we could get that $\lambda_{\max}(R^{\top}RB^{-1}S^{\top}S-B^{-1}) \leq \eps_{0} \kappa \lambda_{\min}(B^{-1})$. Then, we could the following derivation process:
\begin{align*}
0 &~\geq \lambda_{\max}(R^{\top}RB^{-1}S^{\top}S-B^{-1}) - \eps_{0} \lambda_{\min}(B^{-1}) \\
&~\geq \lambda_{\max}(R^{\top}RB^{-1}S^{\top}S - (1+\eps_{0})B^{-1})
\end{align*}

where the first step holds, because of we use Lemma~\ref{lem:CE_two_sketch} to obtain the intermediate result. 
And the second step holds due to the properties of eigenvalue of the matrix. Finally, we could obtain that
\begin{align*}
    R^{\top}RB^{-1}S^{\top}S \preceq (1+\eps_{0})B^{-1}
\end{align*}

\textbf{Case 2:} 

From 
\begin{align*}
|x^{\top}R^{\top}RB^{-1}S^{\top}Sx - x^{\top}B^{-1}x| \leq \eps_{0} \lambda_{\min}(B^{-1}),
\end{align*}
we could get that
\begin{align*}
\lambda_{\min}(R^{\top}RB^{-1}S^{\top}S - B^{-1}) \geq - \eps_{0} \kappa \lambda_{\min}(B^{-1}).
\end{align*}

Then, we could the following equation:
\begin{align*}
0 &~\leq \lambda_{\min}(R^{\top}RB^{-1}S^{\top}S - B^{-1}) + \eps_{0} \lambda_{\min}(B^{-1}) \\
&~\leq \lambda_{\min}(R^{\top}RB^{-1}S^{\top}S - (1-\eps_{0})B^{-1})
\end{align*}
where the first step holds, because we use Lemma~\ref{lem:CE_two_sketch} to obtain the intermediate result. The second step holds because of the properties of eigenvalue.

Finally, according to the above equation, we could obtain that
\begin{align*}
(1 - \eps_{0}) B^{-1} \preceq R^{\top}RB^{-1}S^{\top}S
\end{align*}

Combining two above results, due to the reason that $\eps_{0} = O(\sqrt{n} \log^{3} d/b_{\min}) \kappa$, we could get that
\begin{align*}
(1 - \eps_{0}) B^{-1} \preceq R^{\top}RB^{-1}S^{\top}S \preceq (1 + \eps_{0}) B^{-1}
\end{align*}
for any vector $x \in \R^{n}$ and $\|x\|_{2}=1$. Finally, we could choose $b_{\min} = \eps^{-1} \sqrt{n} \kappa^{2} \log^{3} d$, where $\eps \in (0, 1/10)$, to make $\eps_{0} \in (0, 1/10)$. This finishes the proof.
\end{proof}

\subsection{Bounding \texorpdfstring{$|g^{\top} \wt{P} h - g^{\top} \wh{P} h|$}{}} \label{sec:bound_wt_P_wh_P}

We show that $|g^{\top} \wt{P} h - g^{\top} \wh{P} h|$ can be bounded.

\begin{lemma}[$\wt{P}$ and $\wh{P}$ are close]
If the following conditions hold
\begin{itemize}
    \item $\wt{B}=(R_{1}^{\top}R_{1}AWA^{\top}R_{2}^{\top}R_{2})^{-1}$.
    \item $C=W^{1/2}A^{\top}$.
    \item Let $g \in \R^{n}$ and $h \in \R^{n}$ be two vectors.
    \item Let $R_{1} \in \R^{b_{1} \times d}$, $R_{2} \in \R^{b_{2} \times d}$ are two sketching matrices.
    \item Let $b_{\min} = \{b_{1}, b_{2}\}$.
\end{itemize}
Then,  We have
\begin{align*}
&~|g^\top \wt{P} h - g^\top \wh{P} h| \\
\lesssim &~ \frac{\log^{1.5} d}{\sqrt{b_{\min}}} \cdot (\|g^{\top} C \|_2 \|\wt{B} C^{\top} h\|_2 + \|g^{\top} C \wt{B} \|_2 \|C^{\top} h\|_2) \\
+ &~ \frac{\log^{3} d}{b_{\min}} \cdot \|g^{\top}C\|_2 \|C^{\top} h\|_2 \|\wt{B}\|_F
\end{align*}
with probability at least $1 - 1/\poly(n)$.
\end{lemma}
\begin{proof}
We could using Lemma~\ref{lem:bound_out_inverse} to prove the above lemma. By setting $\wt{B}=(R_{1}^{\top}R_{1}AWA^{\top}R_{2}^{\top}R_{2})^{-1}$, and $C=W^{1/2}A^{\top}$ where $R_{1} \in \R^{b_{1} \times d}$ and $R_{2} \in \R^{b_{2} \times d}$ are two sketching matrices.
\end{proof}

\subsection{Tools for Bounding \texorpdfstring{$|g^{\top} \wt{P} h - g^{\top} \wh{P} h|$}{}} 

We present the tools for bounding $|g^{\top} \wt{P} h - g^{\top} \wh{P} h|$.

\label{sec:tool_bound_wt_P_wh_P}
\begin{lemma}[Tools for showing $\wt{P}$ and $\wh{P}$ are close] \label{lem:bound_out_inverse}
If the following conditions hold
\begin{itemize}
    \item Let $\wt{B} \in \R^{d \times d}$ and $C \in \R^{n \times d}$ be two matrices.
    \item $R \in \R^{b_{1} \times d}$, $S \in \R^{b_{2} \times d}$ are defined as in Definition~\ref{def:ams_matrix}.
    \item $g, h \in \R^{n}$ are vectors.
    \item Let $b_{\min} = \{b_{1}, b_{2}\}$.
\end{itemize}

Then, we have
\begin{align*}
&~g^{\top}C(R^{\top}R)\wt{B}(S^{\top}S)C^{\top}h - g^{\top}C\wt{B}C^{\top}h \\
\lesssim &~ \frac{\log^{1.5} d}{\sqrt{b_{\min}}} \cdot (\|g^{\top} C \|_2 \|\wt{B} C^{\top} h\|_2 + \|g^{\top} C \wt{B}\|_2 \|C^{\top} h\|_2) \\
+ &~ \frac{\log^{3} d}{b_{\min}} \cdot \|g^{\top}C\|_2 \|C^{\top} h\|_2 \|\wt{B}\|_F
\end{align*}
with probability at least $1-1/\poly(n)$.
\end{lemma}

\begin{proof}
This can be proved by using Lemma~\ref{lem:CE_two_sketch}. 
\end{proof}

\subsection{Bounding \texorpdfstring{$|g^{\top} P h - g^{\top} \wt{P} h|$}{}} \label{sec:bound_P_wt_P}

We show that $|g^{\top} P h - g^{\top} \wt{P} h|$ can be bounded.

\begin{lemma}[$P$ and $\wt{P}$ are close]\label{lem:bound_P_wt_P}
If the following conditions hold
\begin{itemize}
    \item Given $A \in \R^{d \times n}$ and $W \in \R^{n \times n}$.
    \item Let $R_{1} \in \R^{b_{1} \times d}$, $R_{2} \in \R^{b_{2} \times d}$, $R_{3} \in \R^{b_{3} \times d}$, and $R_{4} \in \R^{b_{4} \times d}$ be four matrices, defined as in Definition~\ref{def:ams_matrix}.
    \item Let $g \in \R^{n}$ and $h \in \R^{n}$ be two vectors.
    \item Let $P$ be defined as Eq.~(\ref{eq:def_P}), $\wh{P}$ and $\wt{P}$ be defined as Def.~\ref{def:app_P}.
    \item Let $b_{\min}=\min \{b_{1}, b_{2}\}$.
\end{itemize}

Then, we have that
\begin{align*}
|g^{\top} P h - g^{\top} \wt{P} h| 
\lesssim & ~ \log^{6} d \cdot (\frac{1}{\sqrt{b_{\min}}} + \frac{n}{b^{2}_{\min}}) \kappa \|g^{\top}C\|_{2} \|C^{\top} h\|_{2} \|B\|_{2}
\end{align*}
with probability at least $1-1/\poly(n)$, where $C=W^{1/2}A^{\top}$, $B=(AWA^{\top})^{-1}$, and  $\kappa = \lambda_{\max}(B) / \lambda_{\min}(B)$.
\end{lemma}
\begin{proof}
In order to simplify the proof, we first define $B$ as follows: 
\begin{align*}
\wt{B}:=(R_{1}^{\top}R_{1}AWA^{\top}R_{2}^{\top}R_{2})^{-1}.
\end{align*}
We define $C$ as follows:
\begin{align*}
C := W^{1/2}A^{\top} .
\end{align*}
We define $B$ as follows:
\begin{align*}
B := (AWA^{\top})^{-1}.
\end{align*}

By using triangle inequality, we could obtain that
\begin{align*}
|g^{\top} P h - g^{\top} \wt{P} h| \leq |g^{\top} P h - g^{\top} \wh{P} h| + |g^{\top} \wh{P} h - g^{\top} \wt{P} h|.
\end{align*}
By using Lemma~\ref{lem:bound_inverse} and Lemma~\ref{lem:bound_out_inverse}, we could obtain that
\begin{align*}
|g^{\top} P h - g^{\top} \wh{P} h| \leq 2 \eps_{0} \|g^{\top}C\|_{2} \|C^{\top}h\|_{2} \|B\|_{2}
\end{align*}
and
\begin{align*}
&~ |g^{\top} \wh{P} h - g^{\top} \wt{P} h| \\
\lesssim &~ \frac{\log^{1.5} d}{\sqrt{b_{\min}}} \cdot (\|g^{\top} C\|_2 \|\wt{B} C^{\top} h\|_2  + \|g^{\top} C \wt{B} \|_2 \|C^{\top} h\|_2) 
+ ~ \frac{\log^{3} d}{b_{\min}} \cdot \|g^{\top} C \|_2 \|C^{\top} h\|_2 \|\wt{B}\|_F \\
\end{align*}

According to some facts, we could get that $\|AB\|_{2} \leq \|A\|_{2} \cdot \|B\|_{2}$ and $\|A\|_{2} \leq \|A\|_{F} \leq \sqrt{n} \|A\|_{2}$, for $A \in \R^{m \times n}$.

Then, we could get that
\begin{align*}
&~ |g^{\top} P h - g^{\top} \wt{P} h| \\
\leq & ~ |g^{\top} P h - g^{\top} \wh{P} h| + |g^{\top} \wh{P} h - g^{\top} \wt{P} h| \\
\lesssim & ~ 2 \eps_{0} \|g^{\top} C\|_{2} \|C^{\top} h\|_{2} \|B\|_{2} \\
+ & ~ \frac{\log^{1.5} d}{\sqrt{b_{\min}}} \cdot (\|g^{\top} C\|_2 \|\wt{B} C^{\top} h\|_2 + \|g^{\top} C \wt{B} \|_2 \|C^{\top} h\|_2) \\
+ & ~ \frac{\log^{3} d}{b_{\min}} \cdot \|g^{\top} C \|_2 \|C^{\top} h\|_2 \|\wt{B}\|_F \\
\lesssim & ~ 2 \eps_{0} \|g^{\top} C\|_{2} \|C^{\top} h\|_{2} \|B\|_{2}  \\
+ & ~ \frac{\log^{1.5} d}{\sqrt{b_{\min}}} \cdot (1+2\eps_{0}) \|g^{\top} C\|_2 \|C^{\top} h\|_2 \|B\|_{2} \\
+ & ~ \frac{\log^{3} d}{b_{\min}} \cdot (1+2\eps_{0}) \sqrt{n} \cdot \|g^{\top} C \|_2 \|C^{\top} h\|_2 \|B\|_2 \\
\lesssim & ~ (\frac{\log^{1.5}d}{\sqrt{b_{\min}}} + \frac{\log^{3}d}{b_{\min}} + \frac{\sqrt{n} \log^{4.5}d}{b_{\min}^{1.5}} + \frac{n \log^{6}d}{b_{\min}^{2}}) \cdot ~\kappa \|g^{\top}C\|_{2} \|C^{\top} h\|_{2} \|B\|_{2} \\
\lesssim & ~ \log^{6} d \cdot (\frac{1}{\sqrt{b_{\min}}} + \frac{n}{b^{2}_{\min}}) \cdot \kappa \|g^{\top}C\|_{2} \|C^{\top} h\|_{2} \|B\|_{2}
\end{align*}
where the first step derives from the triangle inequality and the second step is due to Lemma~\ref{lem:bound_inverse} and Lemma~\ref{lem:bound_out_inverse}. The third step comes from $\|\wt{B}\|_{F} \leq \sqrt{n} \|\wt{B}\|_{2} \leq (1+\eps_{0}) \sqrt{n} \|B\|_{2}$.

Next, we show the reason that the fourth step holds
\begin{align*}
    & ~ \frac{\log^{1.5}d}{\sqrt{b_{\min}}} + \frac{\log^{3}d}{b_{\min}} + \frac{\sqrt{n} \log^{4.5}d}{b_{\min}^{1.5}} + \frac{n \log^{6}d}{b_{\min}^{2}} \\
    \lesssim & ~ \log^{6} d \cdot (\frac{1}{\sqrt{b_{\min}}} + \frac{1}{b_{\min}} + \frac{\sqrt{n}}{b^{1.5}_{\min}} + \frac{n}{b^{2}_{\min}}) \\
    \lesssim & ~ \log^{6} d \cdot (\frac{1}{\sqrt{b_{\min}}} + \frac{n}{b^{2}_{\min}})
\end{align*}
where the first step follows from $\log^{6} d$ is the dominate item in the numerator, and the second step follows from  $1/\sqrt{b_{\min}} > 1 / b_{\min}, \forall b_{\min} \geq 1$ and $\sqrt{n} / b^{1.5}_{\min} < n / b^{2}_{\min}, \forall b_{\min}\leq n$.
\end{proof}

\section{Main Result} \label{sec:proof_main_res}

In this section, we state the main result of this paper. Next, we give the proof of this statement.

\begin{theorem}[Formal Main Result] \label{thm:main_result_formal}
If the following conditions hold
\begin{itemize}
    \item $\min_{Ax=b, x \in \Pi_{i=1}^{m}K_{i}}c^{\top}x$ is a convex problem under the federated learning setting, where $K_{i}$ is compact convex sets.
    \item For each $i \in [m]$, we are given a $\nu_{i}$-self concordant barrier function $\phi_{i}$ for $K_{i}$. 
    \item We have $x^{(0)}=\arg\min_{x}\sum_{i=1}^m \phi_{i}(x_{i})$.
    \item For all $x \in \prod_{i=1}^m K_{i}$, we have that $\|x\|_{2}$ is bounded by $R$ (Diameter of the set).
    \item $\|c\|_{2}$ is bounded by $L$ (Lipschitz constant of the program).
\end{itemize}

Then, there exists a federated learning algorithm (see Algorithm~\ref{alg:erm_fl}) that runs in $O(\sqrt{\nu} \log^{2} m \log(\frac{\nu}{\delta}))$ iterations and each iteration sends $O(bn)$ words to find a vector $x$ such that
\begin{align*}
    c^{\top}x &~\leq \min_{Ax=b, x\in\Pi_{i=1}^{m}K_{i}} c^{\top}x+LR \cdot \delta \\
    \|Ax-b\|_{1} &~\leq 3 \delta (R \sum_{i=1}^{d} \sum_{j=1}^{n} |A_{i,j}| + \|b\|_{1})
\end{align*}
where $\|c\|_{2} \leq L$, $\|x\|_{2} \leq R$, and $\nu = \sum_{i=1}^m \nu_i$.
\end{theorem}

\begin{proof}
By combining Lemma~\ref{lem:bound_Phi}, Lemma~\ref{lem:feasible_LP} and Lemma~\ref{lem:apx_center_imply_gap}, we could get that a vector $x$ which satisfies the above conditions after $O(\sqrt{\nu} \log^{2} m \log(\frac{\nu}{\delta}))$ iterations. In addition, the Algorithm~\ref{alg:erm_fl} sends $O(bn)$ words at each iteration (Line $14$ in Algorithm~\ref{alg:erm_fl}). This finishes the proof.
\end{proof}

\section{Central Path}
\label{sec:central_path}

Here, we introduce some basic result of central path in Algorithm~\ref{alg:erm_fl}, which could be used to prove the guarantee of $W$ and the main result of this paper. Central path algorithm is a very standard method for solving linear programming \cite{cls19,lsz19,song19,b20,dly21,jswz21,sy21,gs22,qszz23}, semi-definite programming \cite{jklps20,hjs+22,hjs+22_quantum,gs22}. 

We first give the definition of some parameters here:
\begin{definition} \label{def:mu_gamma_c}
For any $i \in [m]$, we let $\phi_{i}(x)$ be defined as in Definition~\ref{def:phi} and let $\mu_{i}^{t}(x,s) \in \R^{n_{i}}$, $\gamma_{i}^{t}(x,s) \in \R$, and $c_{i}^{t}(x,s) \in \R$ be defined as below:
\begin{align}
    &~\mu_{i}^{t}(x,s) = s_{i} / \wt{t} + \nabla\phi_{i}(x_{i}) \label{eq:def_mu} \\
    &~\gamma_{i}^{t}(x,s) = \|\mu_{i}^{t}(x,s)\|_{\nabla^{2}\phi_{i}(x_{i})^{-1}} \label{eq:def_gamma} \\
    &~c_{i}^{t}(x,s) = \begin{cases}
    \frac{\exp(\lambda\gamma_{i}^{t}(x,s))/\gamma_{i}^{t}(x,s)}{(\sum_{i=1}^{m}\exp(2\lambda\gamma_{i}^{t}(x,s)))^{1/2}} & \text{if }\gamma_{i}^{t}(x,s)\geq96\sqrt{\alpha}\\
    0 & \text{otherwise}
    \end{cases} \label{eq:def_c} 
\end{align}
where $\lambda = O(\log m)$, $\wt{t} = (1 - \xi / \sqrt{\nu})^{t-1}$, and $\xi = O(\log^{-2}(m))$. 
\end{definition}

According to the Definition~\ref{def:mu_gamma_c} and Algorithm~\ref{alg:erm_fl}, we could obtain that
\begin{align}
h_{i}^{t} = -\alpha\cdot c_{i}^{t}(\ov{x},\ov{s})\mu_{i}^{t}(\ov{x},\ov{s}) \label{eq:def_h}
\end{align}
where $\alpha=O(1 / \log^{2} m)$. In addition, we define that
\begin{align*}
\Phi^{t}(x^{t},s^{t}) = \sum_{i=1}^{m} \exp(\lambda \|\mu_{i}^{t}(x^{t},s^{t})\|_{\nabla^{2} \phi_{i}(x^{t}_{i})^{-1}})    
\end{align*}
where $\lambda=O(\log m)$. Then, we could obtain the following lemma.

\begin{lemma}[Bounding $\alpha_{i}$] \label{lem:alpha_i}
If the following conditions hold
\begin{itemize}
    \item $\alpha$ represents the parameter in Algorithm~\ref{alg:erm_fl}.
    \item For any $i$ in $[m]$, we have $\alpha_{i}=\|\delta_{x,i}\|_{\ov{x}_{i}}$.
\end{itemize}

Then, we have
\begin{align*}
    \sum_{i=1}^m \alpha_{i}^{2}\leq4\alpha^{2}.
\end{align*}
\end{lemma}

\begin{proof}

Note that
\begin{align*}
\sum_{i=1}^m \alpha_{i}^{2} = &~ \|\delta_{x}\|_{\ov{x}}^{2} \\
= &~ h^{\top}\wt V^{1/2}(I-\wt P)\wt V^{1/2}\nabla^{2}\phi(\ov{x})\wt V^{1/2}(I-\wt P)\wt V^{1/2}h.
\end{align*}
Due to the reason that
\begin{align*}
(1-2\alpha)(\nabla^{2}\phi_{i}(\ov{x}_{i}))^{-1}\preceq\wt V_{i}\preceq(1+2\alpha)(\nabla^{2}\phi_{i}(\ov{x}_{i}))^{-1}
\end{align*}
we have that
\[
(1-\alpha)(\nabla^{2}\phi(\ov{x}))^{-1}\preceq\wt V\preceq(1+\alpha)(\nabla^{2}\phi(\ov{x}))^{-1}.
\]
Using $\alpha\leq\frac{1}{10000}$, we have that 
\[
\sum_{i=1}^m \alpha_{i}^{2} \leq2h^{\top}\wt V^{1/2}(I-\wt P)(I-\wt P)\wt V^{1/2}h\leq2h^{\top}\wt Vh
\]
where we used that $I-\wt P$ is an orthogonal projection at the end. 

Finally, we note that
\begin{align*}
& ~ h^{\top}\wt Vh \\
\leq & ~ 2 \sum_{i=1}^m \|h^{t}_{i}\|_{\ov{x}_{i}}^{*2} \\
 = & ~ 2\alpha^{2}\sum_{i=1}^m c_{i}^{t}(\ov{x},\ov{s})^{2}\|\mu_{i}^{t}(\ov{x},\ov{s})\|_{\ov{x}_{i}}^{*2}\\
\leq & ~ 2\alpha^{2}\sum_{i=1}^m (\frac{ \exp (2 \lambda \gamma_i^t ( \ov{x}, \ov{s} ) ) /  \gamma_i^t ( \ov{x} , \ov{s} )^2 }{ \sum_{i=1}^m \exp( 2 \lambda \gamma_i^t ( \ov{x}, \ov{s} ) ) }  \|\mu_{i}^{t}(\ov{x},\ov{s})\|_{\ov{x}_{i}}^{*2} ) \\
 = & ~ 2\alpha^{2}\frac{ \sum_{i=1}^m \exp(2\lambda\gamma_{i}^{t}(\ov{x},\ov{s}))}{\sum_{i=1}^m \exp(2\lambda\gamma_{i}^{t}(\ov{x},\ov{s}))} \\
 = & ~ 2\alpha^{2} 
\end{align*}
where the second step is from the definition of $h^{t}_i$ (Eq.~(\ref{eq:def_h})), the third step follows from the definition of $c_i^t$ (Eq.~(\ref{eq:def_c})), the fourth step follows from definition of $\gamma_i^t$ (See Eq.~(\ref{eq:def_gamma})).

Therefore, putting it all together, we can show
\begin{align*}
\sum_{i=1}^m \alpha_i^2 \leq 4 \alpha^2.
\end{align*}
\end{proof}

\begin{lemma}[Lemma A.8 in~\cite{lsz19}] \label{lem:bound_Phi}
If $\Phi^{t}(x^{t},s^{t})\leq80\frac{m}{\alpha}$, then
\begin{align*}
\Phi^{t+1}(x^{t+1},s^{t+1}) 
\leq  \left( 1 - \frac{ \alpha \lambda }{40 \sqrt{m}} \right) \Phi^{t}(x^{t}, s^{t}) + \sqrt{m} \lambda \cdot \exp(192 \lambda \sqrt{\alpha}).
\end{align*}
In particularly, we have $\Phi^{t+1} (x^{t+1}, s^{t+1}) \leq 80 \frac{m}{\alpha}$.
\end{lemma}

\section{Initial Point and Termination Condition}

\label{sec:initial_point_and_termination_condi}

Now, we state some basic results of self-concordance function, which could be used to prove the main result of this paper.

\begin{lemma}[Theorem 4.1.7, Lemma 4.2.4 in \cite{n98}]\label{lem:self_concordant}
Let $\phi$ be any $\nu$-self-concordant barrier. 

Then, for any $x,y\in\mathrm{dom}\phi$,
we have
\begin{align*}
\left\langle \nabla\phi(x),y-x\right\rangle  & \leq\nu,\\
\left\langle \nabla\phi(y)-\nabla\phi(x),y-x\right\rangle  & \geq\frac{\|y-x\|_{x}^{2}}{1+\|y-x\|_{x}}.
\end{align*}
Let $x^{*}=\arg\min_{x}\phi(x)$. For any $x\in\R^{n}$ such that
$\|x-x^{*}\|_{x^{*}}\leq1$, we have that $x\in\mathrm{dom}\phi$.
\[
\|x^{*}-y\|_{x^{*}}\leq\nu+2\sqrt{\nu}.
\]

\end{lemma}

\begin{lemma}[Lemma D.2 in~\cite{lsz19}] \label{lem:feasible_LP}
If the following conditions hold
\begin{itemize}
    \item $\min_{Ax=b,x\in\prod_{i=1}^{m}K_{i}}c^{\top}x$ is a convex problem where for each $i$, $K_{i}$ is a compact convex set.
    \item $\phi_{i}$ is defined as in Definition~\ref{def:phi} for $K_{i}$, where $i$ is in $[m]$. 
    \item We have $x^{(0)}=\arg\min_{x}\sum_{i=1}^{m}\phi_{i}(x_{i})$.
    \item Diameter of the set: For any $x\in\prod_{i=1}^m K_{i}$, we have that $\|x\|_{2}\leq R$.
    \item Lipschitz constant of the program: $\|c\|_{2}\leq L$.
\end{itemize}

Then, the modified program $\min_{\overline{A}\overline{x}=\overline{b},\overline{x}\in \prod_{i=1}^m K_{i}\times\R_{+}}\overline{c}^{\top}\overline{x}$
with 
\[
\overline{A}=[A\ |\ b-Ax^{(0)}],\overline{b}=b\text{, and }\overline{c}=\left[\begin{array}{c}
\frac{\delta}{LR}\cdot c\\
1
\end{array}\right]
\]
satisfies the following, for any $\delta > 0$:
\begin{enumerate}
\item $\overline{x}=\left[\begin{array}{c}
x^{(0)}\\
1
\end{array}\right]$, $\overline{y}=0_{d}$ and $\overline{s}=\left[\begin{array}{c}
\frac{\delta}{LR}\cdot c\\
1
\end{array}\right]$ are feasible primal dual vectors with $\|\overline{s}+\nabla\overline{\phi}(\overline{x})\|_{\overline{x}}^{*}\leq\delta$
where $\overline{\phi}(\overline{x})=\sum_{i=1}^{m}\phi_{i}(\overline{x}_{i})-\log(\overline{x}_{m+1})$.
\item For any $\overline{x}$ such that $\overline{A}\overline{x}=\overline{b},\overline{x}\in\prod_{i=1}^m K_{i}\times\R_{+}$
and $\overline{c}^{\top}\overline{x}\leq\min_{\overline{A}\overline{x}=\overline{b},\overline{x}\in\prod_{i=1}^m K_{i}\times\R_{+}}\overline{c}^{\top}\overline{x}+\delta^{2}$,
the vector $\overline{x}_{1:n}$ ($\overline{x}_{1:n}$ is the first
$n$ coordinates of $\overline{x}$) is an approximate solution to
the original convex program in the following sense 
\begin{align*}
c^{\top}\overline{x}_{1:n} & \leq\min_{Ax=b,x\in \prod_{i=1}^m K_{i}}c^{\top}x+LR\cdot\delta,\\
\|A\overline{x}_{1:n}-b\|_{1} & \leq3\delta\cdot\left(R\sum_{i=1}^{d} \sum_{j=1}^{n} |A_{i,j}|+\|b\|_{1}\right),\\
\overline{x}_{1:n} & \in\prod_{i=1}^m K_{i}.
\end{align*}
\end{enumerate}
\end{lemma}

\begin{lemma}[Lemma D.3 in~\cite{lsz19}]
\label{lem:apx_center_imply_gap}
If the following conditions hold
\begin{itemize}
    \item $\phi_{i}(x_{i})$ is defined as in Definition~\ref{def:phi}.
    \item For any $i\in[m]$, we possess $\frac{s_{i}}{t}+\nabla\phi_{i}(x_{i})=\mu_{i}$, $A^{\top}y+s=c$, and $Ax=b$.
    \item $\|\mu_{i}\|_{x,i}^{*}\leq1$ for all $i$.
\end{itemize}

Then, we have that 
\[
\left\langle c,x\right\rangle \leq\left\langle c,x^{*}\right\rangle +4t\nu
\]
where $x^{*}=\arg\min_{Ax=b,x\in \prod_{i=1}^m K_{i}}c^{\top}x$ and $\nu = \sum_{i=1}^m \nu_{i}$.

\end{lemma}

\ifdefined\isarxiv

\bibliography{ref}
\bibliographystyle{alpha}

\else

\fi

\end{document}